\documentclass[letterpaper]{article}
\usepackage{proceed2e}
\usepackage[margin=1in]{geometry}
\usepackage{natbib}
\usepackage{bm}
\usepackage{amssymb}
\usepackage{amsmath}
\usepackage{mathtools}
\usepackage{amsfonts}
\usepackage{bm} 
\usepackage{graphicx}
\usepackage{rotating}
\usepackage{tikz}
\usepackage{tabularx,hhline}
\usepackage{array}
\usepackage{tabulary}
\newcolumntype{K}[1]{>{\centering\arraybackslash}p{#1}}
\usepackage{bbm}
\usepackage{array}
\usepackage{multirow}
\usepackage{caption}
\usepackage{subcaption}
\usepackage{bbm}
\usepackage{tabularx,booktabs} 
\usepackage{standalone}
\usepackage[acronym,smallcaps,nowarn]{glossaries}
\usepackage[utf8]{inputenc} 
\usepackage[T1]{fontenc}    
\usepackage{hyperref}       
\usepackage{url}            
\usepackage{booktabs}       
\usepackage{amsfonts}       
\usepackage{nicefrac}       
\usepackage{microtype} 
\usepackage{etex}
\usepackage{stfloats}
\usepackage[algoruled]{algorithm2e}
\usepackage{listings}
\usepackage{fancyvrb}
\fvset{fontsize=\normalsize}
\usepackage{enumitem}
\usepackage{mathtools,amssymb,lipsum}
\usepackage{cuted}
\usepackage{float}
\usepackage{csquotes}

\usetikzlibrary{shapes}
\usetikzlibrary{fit}
\usetikzlibrary{chains}
\usetikzlibrary{arrows}

\tikzstyle{latent} = [circle,fill=white,draw=black,inner sep=1pt,
minimum size=22pt, font=\fontsize{12}{10}\selectfont, node distance=1]
\tikzstyle{obs} = [latent,fill=gray!25]
\tikzstyle{const} = [rectangle, inner sep=0pt, node distance=1]
\tikzstyle{factor} = [rectangle, fill=black,minimum size=5pt, inner
sep=0pt, node distance=0.4]
\tikzstyle{det} = [latent, diamond]

\tikzstyle{plate} = [draw, rectangle, rounded corners, fit=#1]
\tikzstyle{wrap} = [inner sep=2pt, fit=#1]

\tikzstyle{newwrap} = [inner sep=0pt, fit=#1]
\tikzstyle{gate} = [draw, rectangle, dashed, fit=#1]

\tikzstyle{caption} = [font=\footnotesize, node distance=0] %
\tikzstyle{plate caption} = [caption, node distance=0, inner sep=0pt,
above right=0pt and 0pt of #1.north east] %
\tikzstyle{new plate caption} = [caption, node distance=0, inner sep=0pt,
above right=0pt and 0pt of #1.south west] %
\tikzstyle{factor caption} = [caption] %
\tikzstyle{every label} += [caption] %

\tikzset{>={triangle 45}}

\newcommand{\factoredge}[4][]{ %
  \foreach \f in {#3} { %
    \foreach \x in {#2} { %
      \path (\x) edge[-,#1] (\f) ; %
    } ;
    \foreach \y in {#4} { %
      \path (\f) edge[-,#1] (\y) ; %
    } ;
  } ;
}

\newcommand{\edge}[3][]{ %
  \foreach \x in {#2} { %
    \foreach \y in {#3} { %
      \path (\x) edge [->,#1] (\y) ;%
    } ;
  } ;
}

\newcommand{\factor}[5][]{ %
  \node[factor, label={[name=#2-caption]#3}, name=#2, #1,
  alias=#2-alias] {} ; %
  \factoredge {#4} {#2-alias} {#5} ; %
}

\newcommand{\plate}[4][]{ %
  \node[newwrap=#3] (#2-newwrap) {}; %
  \node[plate caption=#2-newwrap] (#2-caption) {#4}; %
  \node[plate=(#2-newwrap)(#2-caption), #1] (#2) {}; %
}

\newcommand{\newplate}[4][]{ %
  \node[wrap=#3] (#2-wrap) {}; %
  \node[new plate caption=#2-wrap] (#2-caption) {#4}; %
  \node[plate=(#2-wrap)(#2-caption), #1] (#2) {}; %
}

\newacronym{ADVI}{advi}{automatic differentiation variational inference}
\newacronym{ARD}{ard}{automatic relevance determination}
\newacronym{AUC}{auc}{area under the ROC curve}

\newacronym{BBVI}{bbvi}{black-box variational inference}
\newacronym{BCCA}{bcca}{Bayesian canonical correlation analysis}
\newacronym{BIBFA}{bibfa}{Bayesian inter-battery factor analysis}
\newacronym{BASS}{bass}{Bayesian group factor analysis with structured sparsity}

\newacronym{CTM}{ctm}{correlated topic model}
\newacronym{CVI}{cvi}{collapsed variational inference}
\newacronym{CGS}{cgs}{collapsed Gibbs sampling}
\newacronym{CCA}{cca}{canonical correlation analysis}

\newacronym[\glslongpluralkey={deep exponential families}]{DEF}{def}{deep exponential family}
\newacronym{DMIS}{dmis}{deterministic multiple importance sampling}
\newacronym{DSI}{dsi}{dense stability index}

\newacronym{ELBO}{elbo}{evidence lower bound}
\newacronym{EM}{em}{expectation maximization}

\newacronym{FA}{fa}{Factor analysis}

\newacronym{GNTS}{gn-ts}{gamma-normal time series model}
\newacronym{G-REP}{g-rep}{generalized reparameterization}
\newacronym{GFA}{gfa}{group factor analysis}

\newacronym{HDP}{hdp}{hierarchial Dirichlet process}
\newacronym{HBP}{hbp}{hierarchial beta process}

\newacronym{KL}{kl}{{K}ullback-{L}eibler}

\newacronym{LDA}{lda}{latent {D}irichlet allocation}

\newacronym{MF}{mf}{matrix factorization}
\newacronym{MIS}{mis}{multiple importance sampling}
\newacronym{MATLAB}{matlab}{MATLAB}
\newacronym{MSE}{mse}{mean square error}
\newacronym{MCMC}{mcmc}{Markov chain monte carlo}

\newacronym{NIPS}{nips}{Neural Information Processing Systems}
\newacronym{NGFA}{ngfa}{nonparametric Bayesian group factor analysis}

\newacronym{OBBVI}{o-bbvi}{overdispersed black-box variational inference}

\newacronym{RS-VI}{rsvi}{rejection sampling variational inference}

\newacronym{SVI}{svi}{stochastic variational inference}
\newacronym{SSI}{ssi}{sparse stability index}

\newacronym{VI}{vi}{variational inference}
\newacronym{VGP}{vgp}{variational Gaussian process}

\usepackage{times}

\title{Collapsed Variational Inference for Nonparametric Bayesian Group Factor Analysis}

\author{ {\bf Sikun Yang,\ \ \ Heinz Koeppl}\\
Department of Electrical Engineering and Information Technology \\
Technische Universit\"at Darmstadt\\
64283 Darmstadt, Germany \\
\{sikun.yang, heinz.koeppl\}@bcs.tu-darmstadt.de
}

\begin{document}

\maketitle
\linespread{0.99}
\begin{abstract}
Group factor analysis (GFA) methods have been widely used to infer the common structure and the group-specific signals from multiple related datasets in various fields including systems biology and neuroimaging. To date, most available GFA models require Gibbs sampling or slice sampling to perform inference, which prevents the practical application of GFA to large-scale data. In this paper we present an efficient collapsed variational inference (CVI) algorithm for the nonparametric Bayesian group factor analysis (NGFA) model built upon an hierarchical beta Bernoulli process. Our CVI algorithm proceeds by marginalizing out the group-specific beta process parameters, and then approximating the true posterior in the collapsed space using mean field methods. Experimental results on both synthetic and real-world data demonstrate the effectiveness of our CVI algorithm for the NGFA compared with state-of-the-art GFA methods. 
\end{abstract}
\section{Introduction}
Factor analysis (FA) is a powerful tool widely used to infer low-dimensional structure in multivariate data. More specifically, FA models attempt to 
represent a data matrix $\mathbf{X}\in\mathbb{R}^{N\times D}$ by the product of two matrices plus residual noise as
\begin{equation}
\begin{aligned}
\mathbf{X} = \mathbf{F}\mathbf{G} + \mathbf{E}, \notag
\end{aligned}
\end{equation}
where $\mathbf{F}\in\mathbb{R}^{N\times K}$ denotes the factor score matrix, and $\mathbf{G}\in\mathbb{R}^{K\times D}$ denotes the factor loading matrix; $\mathbf{E}\in\mathbb{R}^{N\times D}$ is the residual noise matrix.
For high-dimensional data, 
FA models imposing sparsity-inducing priors~\citep{West03, Piyushi2008, BPFA2009, NSFA2011} or regularizations~\citep{SPCA2006,PMA2009} over the inferred loading matrices are developed to improve interpretability of the inferred low-dimensional structure. For example, in gene expression analysis, a factor loading matrix characterizing the connections between transcription factors and regulated genes are expected to be sparse~\citep{Carvarlho2008}. 

In many real-world applications, we often deal with multiple related datasets -- each comprising a group of variables -- that need to be factorized in a common subspace. For instance, latent Dirichlet allocation~\citep{LDA} and Poisson factor analysis models~\citep{NBP} have been developed to learn the shared latent topics among multiple related documents. Recently, GFA models~\citep{BGFA2012,SGFA2016} using the automatic relevance determination (ARD) prior have been proposed for drug sensitivity prediction and functional neuroimaging. However, the modeling flexibility achieved by these GFA models comes at a price as their inference usually requires Markov chain Monte Carlo (MCMC) to perform posterior computation, which makes them to scale poorly for large-scale GFA problems. Alternatively, variational Bayesian inference has been shown to be efficient for large-scale data by making an independence assumption among latent variables and parameters~\citep{GMEFVI2008}. However, this strong assumption may lead to very inaccurate results in practical applications, especially for GFA problems where latent variables might be tightly coupled.

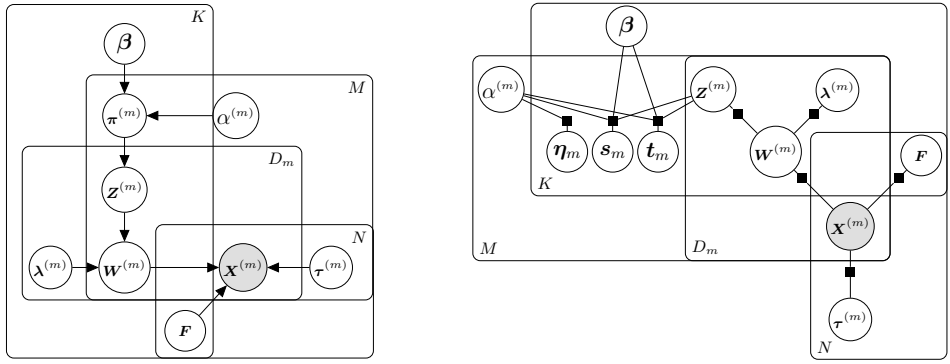
\begin{figure*}[htp!]
  \centering
   \begin{tabular}{c c} 
    \begin{minipage}{.35\textwidth}
      \scalebox{.7}{
      
      \begin{tikzpicture}

  \node[obs]                               (X) {${\scriptstyle \bm{X}}^{\scriptscriptstyle (m)}$};
  \node[latent, below=of X, xshift=-1.1cm, yshift=0.65cm] (F) {${\scriptstyle \bm{F}}$};
  \node[latent, left=of X, xshift=-0.3cm] (G) {${\scriptstyle \bm{W}}^{\scriptscriptstyle (m)}$};
  \node[latent, right=of X, xshift=-0.2cm]  (tau) {${\scriptstyle \bm{\tau}}^{\scriptscriptstyle (m)}$};
  \node[latent, above=of G, yshift=-0.45cm]  (Z) {${\scriptstyle \bm{Z}}^{\scriptscriptstyle (m)}$};
  
  \node[latent, above=of Z, yshift=-0.45cm]  (pi) {${\scriptstyle \bm{\pi}}^{\scriptscriptstyle (m)}$};
  \node[latent, right=of pi, xshift=0.25cm]  (alpha) {${{\alpha}}^{\scriptscriptstyle (m)}$};
  \node[latent, above=of pi, yshift=-0.45cm]  (beta) {${\bm{\beta}}$};
  \node[latent, left=of G, xshift=0.5cm]  (lambda) {${\scriptstyle \bm{\lambda}}^{\scriptscriptstyle (m)}$};

  \edge {G, F, tau} {X} ; 
  \edge {lambda} {G} ;
  \edge {Z} {G} ;
  \edge {pi} {Z} ;
  \edge {beta} {pi} ;
  \edge {alpha} {pi};
  \plate {a} {(Z)(G)(lambda)(X)} {$D_m$} ;
  \plate {b} {(F)(tau)(X), xshift=-0.042cm} {$N$} ;
  \plate {c} {(beta)(pi)(lambda)(Z)(G)(F), xshift=-0.3cm} {$K$} ;
  \plate {d} {(pi)(tau)(Z)(G)(X), xshift=-0.11cm} {$M$} ;

\end{tikzpicture}
      }
    \end{minipage}
    &
      \begin{minipage}{.35\textwidth}
       \scalebox{.7}{
       
       \begin{tikzpicture}
  \node[latent, xshift=0cm]            (tau) {${\scriptstyle \bm{\tau}}^{\scriptscriptstyle (m)}$} ; %
   
  \node[obs, above=0.9of tau, xshift=0cm]            (X) {${\scriptstyle \bm{X}}^{\scriptscriptstyle (m)}$} ; %
  \node[latent, above left=1.05 of X, xshift=0cm]  (G)   {${\scriptstyle \bm{W}}^{\scriptscriptstyle (m)}$}; %
  \node[latent, above left= 0.75 of G, xshift=0cm]  (Z)   {${\scriptstyle \bm{Z}}^{\scriptscriptstyle (m)}$};
  \node[latent, above right=0.75 of G, xshift=0cm]  (lambda)   {${\scriptstyle \bm{\lambda}}^{\scriptscriptstyle (m)}$};
  \node[latent, above right= 1.05of X, xshift=0cm] (F)   {${\scriptstyle \bm{F}}$}; %
  \node[latent, above left=1.5 of Z,yshift=-0.45cm] (beta)   {$\bm{\beta}$}; 
  
  \node[latent, left= 1.35 of G] (t)   {$\bm{t}_m$};
  
  \node[latent,  left= 0.075 of t] (s)   {$\bm{s}_m$};
  \node[latent, left = 0.075 of s] (eta)   {$\bm{\eta}_m$};
  \node[latent, left = 3.05 of Z, xshift=-0.1cm] (alpha)   {${\alpha}^{\scriptscriptstyle (m)}$};
  
  \factor[above=of eta, yshift=-0.3cm] {alpha-eta} { } {alpha} {eta} ; %
  \factor[above=of s, yshift=-0.3cm] {alpha-s} { } {alpha, beta, Z} {s} ; %
  \factor[above=of t, yshift=-0.3cm] {alpha-t} { } {alpha, beta, Z} {t} ; %
  \factor[above left=of G] {z-g} { } {Z} {G} ;
  \factor[above right=of G] {lambda-g} { } {lambda} {G} ;
  \factor[above left=0.7of X] {g-x} { } {G} {X} ;
  \factor[above right=0.7of X] {f-x} { } {F} {X} ;
  \factor[above=of tau] {x-tau} { } {X} {tau} ;
  
  \newplate {} { %
    (Z)(G)(lambda)(X), xshift=0.1cm
  } {$D_m$} ;
  
  \newplate {} { %
    (Z)(G)(lambda)(G)(beta)(eta)(s)(t)(F), yshift=-0.15cm, xshift=-0.1cm
  } {$K$} ;
  
  \newplate {} { %
    (alpha)(alpha-eta)(eta)(s)(t)(Z)(G)(lambda)(X) , xshift=0.1cm
  } {$M$};
   \newplate {} { %
    (tau)(F)(X), yshift=-0.15cm,xshift= -0.1cm
  } {$N$};
\end{tikzpicture}
       
       }
    \end{minipage}
  \end{tabular}
  \caption{Left: The graphical representation of the proposed model. Right: Factor graph of the model with auxiliary variables.}
  \label{gm}
\end{figure*}

Motivated by this limitation, we propose a computationally efficient collapsed variational inference algorithm for the nonparametric Bayesian group factor analysis model. 
Our NGFA model is built upon the hierarchical beta process (HBP)~\citep{HBP}. We note that the HBP has been investigated in~\citep{Chen2011,GibbsHBP,SliceHBP} for joint modeling of multiple data matrices utilizing MCMC, but again showed poor scalability and slow convergence.
For nonparametric Bayesian models, such as HDP topic model~\citep{HDP} and HDP hidden Markov models~\citep{fox2011}, collapsed Gibbs sampling (CGS) are typically employed to perform posterior computation because CGS rapidly convergences onto the true posterior. However, it remains challenging to assess the convergence of CGS algorithms for practical use. To address this issue, collapsed variational inference algorithms~\citep{CVBLDA,CVBHDP,SCVBLDA} are developed for topic models by integrating out model parameters, and then applying the mean field approximation to the latent variables.
%
%
Recently, collapsed variational inference algorithms have been developed for hidden Markov models~\citep{CVBHMM}, nonparametric relational models~\citep{ACVB} and Markov jump processes~\citep{CVBMJP} with encouraging results. In this paper, we aim to develop a collapsed variational inference algorithm for the nonparametric Bayesian group factor analysis model. 

We make the following contributions: 
\begin{itemize}[noitemsep]
\item We tackle the group factor analysis problems using a Bayesian nonparametric method based on the hierarchical beta Bernoulli process. The total number of factors is automatically learned from data. Specifically, the NGFA model induces both group-wise and element-wise structured sparsity effectively compared to state-of-the-art GFA methods (see Section 4.1). 
\item An efficient collapsed variational inference algorithm is proposed to infer the NGFA model. 
 \item We apply the developed method to real world multiple related dataset, with encouraging results (see Section 4.2; 4.3).
 \end{itemize}
The paper is organized as follows. In Section 2, we describe the nonparametric Bayesian group factor analysis model. Our collapsed variational inference algorithm for the NGFA is introduced in Section 3. Experimental results are presented in Section~4. Finally, conclusions and possible directions for future research are discussed in Section 5.

\section{Nonparametric Bayesian Group Factor Analysis}
Given multiple related data matrices $\mathbf{X}^{\scriptscriptstyle(1)}, \mathbf{X}^{\scriptscriptstyle(2)}, \ldots, \mathbf{X}^{\scriptscriptstyle(M)}$, each with $N$ samples, i.e., $\mathbf{X}^{\scriptscriptstyle(m)} \in \mathbb{R}^{N\times D_m}$,
our goal is to factorize each dataset $\mathbf{X}^{\scriptscriptstyle(m)}$ into the product of a common factor matrix $\mathbf{F}=[\mathbf{f}_1,\ldots, \mathbf{f}_K]$ of size $N\times K$, and a group-specific factor loading matrix $\mathbf{G}^{\scriptscriptstyle(m)}$ of size $K\times D_m$ as
\begin{equation}
\begin{aligned}
\mathbf{X}^{\scriptscriptstyle(m)} = \mathbf{F}\mathbf{G}^{\scriptscriptstyle(m)} + \mathbf{E}^{\scriptscriptstyle(m)},\label{likelihood}
\end{aligned}
\end{equation}
where $\mathbf{E}^{\scriptscriptstyle(m)} = [\mathbf{e}_1^{\scriptscriptstyle(m)}, \ldots, \mathbf{e}_{D_m}^{\scriptscriptstyle(m)}]$ is assumed to be Gaussian noise for the $m$-th dataset or group. 
We impose independent normal priors over $\mathbf{e}_d^{\scriptscriptstyle(m)} \in \mathbb{R}^{N}$, i.e., $\mathbf{e}_d^{\scriptscriptstyle(m)} \sim \mathcal{N}(0, \mathrm{diag}(\tau_1^{\scriptscriptstyle(m)}, \ldots, \tau_N^{\scriptscriptstyle(m)}))$, where $\tau_n^{\scriptscriptstyle(m)}$ controls the variance of $N$-th sample in the $m$-th group.
As commonly used in factor analysis~\citep{Piyushi2008, BPFA2009, NSFA2011}, we put a normal prior on each factor $\mathbf{f}_{k}$, i.e., $\mathbf{f}_{k} \sim \mathcal{N}(0, \mathbf{I}_{N})$, where $\mathbf{I}_N$ is an identity matrix of size $N$.
To explicitly capture the sparsity, we model the factor loading matrix $\mathbf{G}^{\scriptscriptstyle(m)}$ for each group by the element-wise product of a {binary} matrix $\mathbf{Z}^{\scriptscriptstyle(m)}$ and a real-valued weight matrix $\mathbf{W}^{\scriptscriptstyle(m)}$, i.e., $\mathbf{G}^{\scriptscriptstyle(m)} = \mathbf{Z}^{\scriptscriptstyle(m)} \odot \mathbf{W}^{\scriptscriptstyle(m)}$. 
More specifically, we place a normal prior over each element of $\mathbf{W}^{\scriptscriptstyle(m)}$, i.e., $w_{kd}^{\scriptscriptstyle(m)}\sim \mathcal{N}(0, (\lambda_{kd}^{\scriptscriptstyle(m)})^{-1})$.
To allow the number of factors $K$ to be automatically inferred from data, we model each row of $\mathbf{Z}^{\scriptscriptstyle(m)}$ as a draw from a group-specific Bernoulli process. As our goal is to factorize multiple related data matrices using a common set of factors, we naturally consider the hierarchical beta process \citep{HBP} that allows us to generate a set of latent factors from a \emph{global} beta process $B$, and then allow the generated factors to be shared among all the groups. The usage of the generated factors in each group is determined by the group-specific beta process $A^{(m)}$. More specifically, the hierarchical beta Bernoulli process is
\begin{align}
B &\equiv \sum_{k=1}^{K} \beta_k \delta_{\mathbf{f}_k},\qquad A^{\scriptscriptstyle(m)} \equiv \sum_{k=1}^{K} \pi_k^{\scriptscriptstyle(m)} \delta_{\mathbf{f}_k}, \label{NGFA}\\
\mathbf{f}_k & \sim \mathcal{N}(0, \mathbf{I}_N),\qquad \beta_k \sim \mathrm{Beta}(\kappa_0/K, \kappa_0(K-1)/K),\notag\\\
\pi_k^{\scriptscriptstyle(m)} &\sim \mathrm{Beta}(\alpha^{\scriptscriptstyle(m)}\beta_k, \alpha^{\scriptscriptstyle(m)}\bar{\beta}_k),\notag\\\
z_{kd}^{\scriptscriptstyle(m)} &\sim\mathrm{Bern}(\pi_k^{\scriptscriptstyle(m)}),\notag\
\end{align}
where $\bar{\beta}_k \equiv 1-\beta_k$, and $K$ is a truncation level that is set sufficiently large to ensure a good  approximation to the truly infinite model. The concentration parameters of the global beta process and the local group-specific beta process are $\kappa_0$ and $\alpha^{\scriptscriptstyle(m)}$, respectively. The total number of factors shared among all groups is determined by $\kappa_0$, and the amount of variability of each $A^{\scriptscriptstyle(m)}$ around $B$ is determined by $\alpha^{\scriptscriptstyle(m)}$. 
\begin{figure*}[htp!]
  \centering
\begin{align}
p(\mathbf{Z} \mid \bm{\beta}, \bm{\alpha}) & = \int p(\mathbf{Z} \mid \bm{\pi})p(\bm{\pi} \mid \bm{\beta}, \bm{\alpha})\mathrm{d}\bm{\pi} =\prod_{m,k} \frac{\Gamma{(\alpha^{\scriptscriptstyle(m)}})}{\Gamma{(\alpha^{\scriptscriptstyle(m)} + D_m})} \frac{\Gamma(\alpha^{\scriptscriptstyle(m)}\beta_k + \hat{n}_{mk})}{\Gamma(\alpha^{\scriptscriptstyle(m)}\beta_k)} \frac{\Gamma(\alpha^{\scriptscriptstyle(m)}\bar{\beta}_k +\tilde{n}_{mk})}{\Gamma(\alpha^{\scriptscriptstyle(m)}\bar{\beta}_k)} \notag
\end{align}
 \caption{The marginal distribution of $\mathbf{Z}$. We define $\hat{n}_{mk} \equiv \sum_{d} \mathbbm{1}(z_{kd}^{{\scriptscriptstyle(m)}}=1)$ and $\tilde{n}_{mk} \equiv \sum_{d} \mathbbm{1}(z_{kd}^{{\scriptscriptstyle(m)}}=0)$, where \\$\mathbbm{1}(\cdot)$ is the standard indicator function.}
\label{eq_1}
\end{figure*}
To improve the flexility of the model, we place gamma priors on $\lambda_{kd}^{\scriptscriptstyle(m)}$, $\tau_n^{\scriptscriptstyle(m)}$ and $ \alpha^{\scriptscriptstyle(m)}$, respectively, as
$
\lambda_{kd}^{\scriptscriptstyle(m)} \sim \mathrm{Gam}(g_0,h_0)$, 
$
\tau_n^{\scriptscriptstyle(m)} \sim \mathrm{Gam}(e_0,f_0)$, 
$\alpha^{\scriptscriptstyle(m)} \sim \mathrm{Gam}(c_0,d_0)$. 
The graphical representation of the NGFA model is shown in shown in Fig.~\ref{gm} (left).

\section{Collapsed Variational Inference}
The main idea of collapsed variational inference is to marginalize out model parameters, and then apply the mean field method to approximate the distribution over latent variables. 
%
%
We note that marginalizing out the parameters induces dependencies among the latent variables.
However, each latent variable interacts with the remaining variables only through the sufficient statistics (i.e. the field) in the collapsed space, and the influence of any single variable on the field is small. Hence, the dependency between any two latent variables is weak, suggesting that the mean field assumption is better justified in the collapsed space.
%
In our case, we first marginalize out the group-specific beta process parameters to obtain the marginal distribution over latent variables. We then employ the variational posterior to approximate the distribution of latent variables and the remaining parameters.
\\
\noindent\textbf{Notation.} When expressing the conditional distribution, we will use the shorthand ``--'' to denote full conditionals, i.e., all other variables. For the sake of clarity, we use $\mathbf{X}$ to denote the set of matrices $\left(\mathbf{X}^{\scriptscriptstyle(1)}, \mathbf{X}^{\scriptscriptstyle(2)}, \ldots, \mathbf{X}^{\scriptscriptstyle(M)}\right)$. Similarly, let $\mathbf{Z}$ denote $\left(\mathbf{Z}^{\scriptscriptstyle(1)}, \ldots, \mathbf{Z}^{\scriptscriptstyle(M)}\right)$, and $\bm{\pi}$ denote $\left(\bm{\pi}^{\scriptscriptstyle(1)}, \ldots, \bm{\pi}^{\scriptscriptstyle(M)}\right)$. With slight notational abuse we use generic $p$ to denote probability density and mass functions. 
\\
We repeatedly exploit the following three results~\citep{CVBHDP} to derive the collapsed variational inference algorithm for the NGFA.
\\
\noindent\textbf{Result 1.}
The geometric expectation of a non-negative random variable $y$ is defined as
$
\mathsf{G}[y] \equiv \exp({\mathsf{E}[{\log(y)}]}).$
If $y$ is gamma distributed, i.e.,  $p( y\mid a, b) \propto y^{a-1}e^{-by}$, the geometric expectation of $y$ is
$
\mathsf{G}[y] = \frac{\exp({\Psi(a)})}{b},
$
where $\Psi(y) = \frac{\partial \log \Gamma(y)}{\partial y}$ is the digamma function.
For a beta distributed random variable $y$, i.e., $p(y\mid a, b) \propto y^{a-1}(1-y)^{b-1}$, the geometric expectation of $y$ is $\mathsf{G}[y] = \frac{\exp[\Psi(a)]}{\exp[\Psi(a+b)]}$.
If $y_1, \ldots, y_K$ are mutually independent, we have, 
$\mathsf{G}\Big[\prod_{k=1}^{K} y_k\Big] = \prod_{k = 1}^{K}\mathsf{G}[y_k]$.
\\
\noindent\textbf{Result 2.}
According to the central limit theorem, if $y$ is the sum of $N$ independent Bernoulli random variables, i.e., $y = \sum_{i=1}^N u_i$, where $u_i\sim\mathrm{Bern}(\xi_i)$, then for large enough $N$, $y$ is well approximated by a Gaussian random variable with mean and variance 
as
\begin{align}
\mathsf{E}\left[ y \right] & = \sum_{i=1}^N \xi_i,\qquad \mathsf{V}\left[ y \right] = \sum_{i=1}^N \xi_i \left(1 - \xi_i\right),\notag
\end{align}
respectively. Moreover, the expectation of $\log(y)$ can be approximated using the second-order Taylor expansion~\citep{delta_method} as
\begin{align}
\mathsf{E}\left[ \log(y) \right] & \approx \log(\mathsf{E}\left[ y \right]) - \frac{\mathsf{V}\left[ y \right]}{2(\mathsf{E}\left[ y \right])^{2}}.\notag
\end{align}
\\
\noindent\textbf{Result 3.}
If $l$ is the sum of independent Bernoulli random variables, i.e., $l = \sum_i u_i$, where $u_i\sim\mathrm{Bern}(\xi_i)$, we use
$p_{\scalebox{0.8}{+}}(l)$ to denote the probability of $l$ being positive, i.e., 
\begin{align}
p_{\scalebox{0.8}{+}}(l) &\equiv p(l > 0) =
1 - \prod_i p(u_i=0) \notag\\
&= 1 - \exp\left[\sum_i\log(1 - \xi_i)\right].\notag
\end{align}
Accordingly, the expectation and variance conditional on $l>0$ are defined as $\mathsf{E}_{\scalebox{0.8}{+}}[l] \equiv \frac{\mathsf{E}[l]}{p_{\scalebox{0.8}{+}}(l)}$
and $\mathsf{V}_{\scalebox{0.8}{+}}[l] \equiv \frac{\mathsf{V}[l]}{p_{\scalebox{0.8}{+}}(l)}$, respectively. If $y$ is then a Chinese restaurant table (CRT)~\citep{csp} distributed random variable, i.e., $p(y\mid a, l) = \frac{\Gamma(a)}{\Gamma(a+l)}{ l \brack y}a^y$, where $y = 0,1,\ldots, l$, and ${n \brack m}$ denoting the unsigned Stirling number of the first kind, then the expectation of $y$ can be closely approximated using the improved second-order Taylor expansion as
\begin{align}
&\mathsf{E}[y] \approx \mathsf{G}[a] {p}_{\scalebox{0.8}{+}}(l) \Big( \Psi\big( \mathsf{G}[a]+\mathsf{E}_{\scalebox{0.8}{+}}[l] \big) \notag \\
&- \Psi( \mathsf{G}[a] ) + \frac{{\mathsf{V}_{\scalebox{0.8}{+}}[l] \Psi'( \mathsf{G}[a]+\mathsf{E}_{\scalebox{0.8}{+}}[l]})}{2} \Big),\notag
\end{align}
where $\Psi'(y) = \frac{\partial^2 \log \Gamma(y)}{\partial y^2}$ is the trigamma function. 
\subsection{Collapsed representation}
First, we describe how to obtain the marginal distribution of latent variables. In the next subsection, we will then describe how to derive the CVI algorithm in the collapsed space.

For the NGFA introduced in the previous section, integrating out $\bm{\pi}$ yields the marginal distribution of $\mathbf{Z}$ shown in Fig.~\ref{eq_1} because beta priors are conjugate to Bernoulli distributions. 
As the ratios of gamma functions in Fig.~\ref{eq_1} give rise to difficulties for updating hyperparameter posteriors, we augment the marginal distribution $\mathbf{Z}$ by introducing three sets of auxiliary variables. More specifically, using the auxiliary variable method~\citep{HDP}, the first ratio of gamma function can be re-expressed as
\begin{align}
&\frac{\Gamma{(\alpha^{\scriptscriptstyle(m)}})}{\Gamma{(\alpha^{\scriptscriptstyle(m)} + D_{{\scriptscriptstyle m}}})} \label{gam_1}\\
&= \frac{1}{\Gamma(D_{\scriptscriptstyle m})}\int_{0}^{1} \eta_m^{\alpha^{\scriptscriptstyle(m)}} (1-\eta_m)^{D_{\scriptscriptstyle m}-1} \left(1+\frac{D_{\scriptscriptstyle m}}{\alpha^{\scriptscriptstyle(m)}}\right) \mathrm{d} \eta_{\scriptscriptstyle m}.\notag
\end{align} 
Via the relation between the gamma function and the Stirling numbers of the first kind~\citep{HDP}, the second and third ratio of gamma functions can be re-expressed, respectively, as
\begin{align}
&\frac{\Gamma(\alpha^{\scriptscriptstyle(m)}\beta_k + \hat{n}_{mk})}{\Gamma(\alpha^{\scriptscriptstyle(m)}\beta_k)}
 = \sum_{s_{mk} = 0}^{\hat{n}_{mk}} {\hat{n}_{mk} \brack s_{mk}} (\alpha^{\scriptscriptstyle(m)}\beta_k)^{s_{mk}}, \label{gam_2} \\
 &\frac{\Gamma(\alpha^{\scriptscriptstyle(m)}\bar{\beta}_k +\tilde{n}_{mk})}{\Gamma(\alpha^{\scriptscriptstyle(m)})} 
=\sum_{t_{mk} = 0}^{\tilde{n}_{mk}}{\tilde{n}_{mk} \brack t_{mk}} \big(\alpha^{\scriptscriptstyle(m)}\bar{\beta}_k\big)^{t_{mk}}. \label{gam_3}
\end{align}

Substituting (Eqs.~\ref{gam_1};~\ref{gam_2};~\ref{gam_3}) into Fig.~\ref{eq_1}, we immediately obtain the joint distribution of the latent and auxiliary variables as
\begin{align}
p(\mathbf{Z}, \mathbf{s}, \mathbf{t}, \bm{\eta} \mid \bm{\beta}, \bm{\alpha}) 
&\propto \prod_{m,k} 
\eta_m^{\alpha^{\scriptscriptstyle(m)}-1} (1-\eta_m)^{D_m-1} 
\label{eq_3}\\
 &\times
 { \hat{n}_{mk} \brack s_{mk}} (\alpha^{\scriptscriptstyle(m)}\beta_k)^{s_{mk}}
{ \tilde{n}_{mk} \brack t_{mk}} \big(\alpha^{\scriptscriptstyle(m)}\bar{\beta}_k\big)^{t_{mk}}. \notag
 \end{align}
The factor graph of the expanded system with auxiliary variables is shown in Fig.~\ref{gm} (right).
The conditional distribution of a single latent variable $z_{kd}^{\scriptscriptstyle(m)}$ can be derived using the marginal distribution of $\mathbf{Z}$ and the likelihood function according to Eq.~\ref{likelihood} as
\begin{align}
&p(z_{kd}^{\scriptscriptstyle(m)} = 1 \mid -) \propto \exp\left[\log(\alpha^{\scriptscriptstyle(m)}\beta_k + \hat{n}_{km}^{\neg d})\right] 
\label{p_z_kd}
\\
&\times
\exp\left[-\frac{1}{2}\sum_n \tau_n^{\scriptscriptstyle(m)}\left(\left(w_{kd}^{\scriptscriptstyle(m)}\right)^{2} f_{nk}^2 - 2 w_{kd}^{\scriptscriptstyle(m)} \left.\tilde{x}_{nd}^{\scriptscriptstyle(m)}\right.^{\neg k}\right)\right],\notag
\end{align}
where $\left.(\tilde{x}_{nd}^{\scriptscriptstyle(m)})\right.^{\scriptscriptstyle \neg k} \equiv \big(x_{nd}^{\scriptscriptstyle(m)} - \sum_{j \neq k} z_{jd}^{\scriptscriptstyle(m)}w_{jd}^{\scriptscriptstyle(m)}f_{nj}\big)$, and $\hat{n}_{km}^{\neg d}\equiv \sum_{d'\neq d}\mathbbm{1}(z_{kd'}^{\scriptscriptstyle(m)}=1)$. 
\subsection{Variational approximation}
\begin{figure*}[htp!]
  \centering
\begin{align}
q(z_{kd}^{\scriptscriptstyle(m)}=1) \propto \exp\Bigg\{ \mathsf{E}\left[ \log\left(\alpha^{\scriptscriptstyle(m)}\beta_k + {\hat{n}_{mk}}^{\neg d}\right) \right]  - \frac{1}{2} \sum_n \mathsf{E}\big[\tau_n^{\scriptscriptstyle(m)}\big]\left(%
\mathsf{E}\big[ \big( w_{kd}^{\scriptscriptstyle(m)} \big)^{\scriptscriptstyle2}\big]
\mathsf{E}\big[ f_{nk}^2 \big] 
- 2 \mathsf{E}\big[ w_{kd}^{\scriptscriptstyle(m)} \big] \mathsf{E}\big[ f_{nk} \big] \left.\tilde{x}_{nd}^{\scriptscriptstyle(m)}\right.^{\scriptscriptstyle \neg k}%
 \right) \Bigg\} \label{rho_kd}
\end{align}
\vspace{-1em}
\caption{The variational update for each latent variable.}
\label{eq_5}
\end{figure*}
Next, we shall introduce the variational approximation for our expanded system. For the sake of simplicity, the remaining parameters $(\mathbf{W}, \mathbf{F}, \bm{\beta}, \bm{\lambda}, \bm{\tau}, \bm{\alpha})$ is denoted by $\bm{\theta}$. Formally, the variational posterior over the augmented variables system is assumed to be of the form
\begin{align}
& q(\mathbf{Z}, \bm{\theta}, \mathbf{s}, \mathbf{t}, \bm{\eta}) = q(\bm{\theta}) q(\mathbf{s},\mathbf{t},\bm{\eta} \ |\ \mathbf{Z}) q(\mathbf{Z}),\notag
\end{align}
where $q(\bm{\theta}) \equiv q(\mathbf{W}) q(\mathbf{F}) q(\bm{\beta}) q(\bm{\lambda}) q(\bm{\tau}) q(\bm{\alpha})$.
Note that the true posterior $p(\mathbf{s},\mathbf{t},\bm{\eta} \mid \mathbf{Z})$ is used in our variational update subsequently. 
\\
\noindent\textbf{Evidence Lower Bound (ELBO):} The log marginal likelihood of data is lower bounded as
\begin{align}
\log p(\mathbf{X}\mid \kappa_0) 
&\geq \mathcal{L}(q(\bm{\theta}) q(\mathbf{s},\mathbf{t},\bm{\eta} \ |\ \mathbf{Z}) q(\mathbf{Z}))
\label{ELBO}\\
& = \mathsf{E}_{q(\bm{\theta},\mathbf{Z})} \left[ \log p(\mathbf{X}, \mathbf{Z}, \bm{\theta}\mid  \kappa_0) -q(\bm{\theta},\mathbf{Z}) \right]. \notag
\end{align}
See the supplementary material (A.3) for details.

To maximize the ELBO in Eq.~\ref{ELBO} with respect to the variational parameters, we can take the gradients of the ELBO w.r.t. each parameter, and set it equal to zero. Then, our CVI algorithm proceeds by updating the variational parameters in a coordinate-wise manner. 
\\
\noindent\textbf{Updating $q(\mathbf{Z})$:} The variational update for each latent variable $z_{kd}^{\scriptscriptstyle(m)}$ is
\begin{align}
&q(z_{kd}^{\scriptscriptstyle(m)}=1) \propto \exp\left(\mathsf{E}_{q(\mathbf{Z},\bm{\theta}\ \backslash \ z_{kd}^{\scriptscriptstyle(m)})}\left[\log p(\mathbf{X}, \mathbf{Z}, \bm{\theta}\mid \kappa_0)\right]\right)\notag\\
&\propto \exp\left(\mathsf{E}_{q(\mathbf{Z},\bm{\theta}\ \backslash \ z_{kd}^{\scriptscriptstyle(m)})}\left[\log p(z_{kd}^{\scriptscriptstyle(m)}=1\mid -)\right]\right),\label{z_kd}
\end{align}
where $(\mathbf{Z},\bm{\theta}\ \backslash \ z_{kd}^{\scriptscriptstyle(m)})$ means all the variables and parameters excluding $z_{kd}^{\scriptscriptstyle(m)}$.

Plugging Eq.~\ref{p_z_kd} into Eq.~\ref{z_kd}, we obtain the variational update for $q(z_{kd}^{\scriptscriptstyle(m)}=1)$ in Fig.~\ref{eq_5}.
The exact computation of the log count in Fig.~\ref{eq_5} is too expensive in practice. According to Result 2, we can approximate it as
\begin{align}
\mathsf{E}\left[ \log\left(\alpha^{\scriptscriptstyle(m)}\beta_k + {\hat{n}_{mk}}^{\neg d}\right) \right] \approx \log\left( \mathsf{G}[\alpha^{\scriptscriptstyle(m)}\beta_k] + \mathsf{E}\left[ \hat{n}_{mk}^{\neg d} \right] \right) \notag\\
- \frac{\mathsf{V}\left[ \hat{n}_{mk}^{\neg d} \right] }{2\left( \mathsf{G}[\alpha^{\scriptscriptstyle(m)}\beta_k] + \mathsf{E}\left[ \hat{n}_{mk}^{\neg d} \right] \right)^2},\notag
\end{align}
where the mean and variance of ${\hat{n}_{mk}}^{\neg d}$ are given by
\begin{align}
\mathsf{E}\left[ {\hat{n}_{mk}}^{\neg d} \right] & = \sum_{d'\neq d} q(z_{kd}^{\scriptscriptstyle(m)}=1), \notag\\
\mathsf{V}\left[ {\hat{n}_{mk}}^{\neg d} \right] &= \sum_{d'\neq d} q(z_{kd}^{\scriptscriptstyle(m)}=1) q(z_{kd}^{\scriptscriptstyle(m)}=0).\notag
\end{align}
\\
\noindent\textbf{Updating auxiliary variables:} 
Now we explain how to update the auxiliary variables efficiently using Gaussian approximation techniques.
The variational posteriors for the auxiliary variables $\bm{\eta}$ is
\begin{alignat}{2}
& q(\bm{\eta} \mid \mathbf{Z})  && \propto \prod_{m} \eta_m^{\mathsf{E}[\alpha^{\scriptscriptstyle(m)}] - 1} (1-\eta_m)^{D_m-1}.\notag
\end{alignat}
As $\bm{\eta}$ is beta distributed, via the geometric expectation of Result 1, we have
\begin{align}
\mathsf{E}[\log(\eta_m)] &= \log\left[\mathsf{G}(\eta_m)\right] 
= \Psi(\mathsf{E}[\alpha^{\scriptscriptstyle(m)}]) - \Psi(\mathsf{E}[\alpha^{\scriptscriptstyle(m)}] + D_m). \notag
\end{align}
The variational posteriors for the auxiliary variables $\mathbf{s}$ is
\begin{alignat}{2}
& q(\mathbf{s} \mid \mathbf{Z})  && \propto \prod_{m,k} { \hat{n}_{mk} \brack {s}_{mk}} (\mathsf{G}[\alpha^{\scriptscriptstyle(m)}\beta_k])^{{s}_{mk}}, \label{s} 
\end{alignat}
where the expectation of $\mathbf{s}$ depends on $\mathbf{Z}$ through the count $\hat{n}_{mk}$ that can take many values. Hence, the exact computation of Eq.~\ref{s} is too expensive. According to Result 3, we use the improved second-order Taylor expansion to approximate the expectation of $s_{\scriptscriptstyle mk}$ as
\begin{align}
&\mathsf{E}[s_{\scriptscriptstyle mk}] \approx \mathsf{G}[\alpha^{\scriptscriptstyle(m)}\beta_k] p_{\scalebox{0.8}{+}}( \hat{n}_{mk}) \Big( \Psi\big( \mathsf{G}[\alpha^{\scriptscriptstyle(m)}\beta_k]+\mathsf{E}_{\scalebox{0.8}{+}}[ \hat{n}_{mk}] \big) \notag \\
&-\Psi( \mathsf{G}[\alpha^{\scriptscriptstyle(m)}\beta_k] ) + \frac{{\mathsf{V}_{\scalebox{0.8}{+}}[ \hat{n}_{mk}] \Psi'( \mathsf{G}[\alpha^{\scriptscriptstyle(m)}\beta_k]+\mathsf{E}_{\scalebox{0.8}{+}}[ \hat{n}_{mk}]})}{2} \Big).\notag
\end{align}
Likewise, we can derive the variational update for $\mathbf{t}$ in the same manner. 
Following the exponential family computation~\citep{GMEFVI2008}, the variational updates for the remaining parameters are obtained via the conjugacy of our model specification. We present these variational updates in the supplementary material (A.4).

\section{Experiments}

In this section, we compare the nonparametric Bayesian group factor analysis using our proposed CVI algorithm with the state-of-the-art GFA models. We evaluate the proposed CVI algorithm on both synthetic data and real-world applications. 
In all our experiments, we set $\kappa_0 = 1, c_0=0.1, d_0=0.1, g_0=0.1, h_0=0.1, e_0=0.1, f_0=0.1$. Similar results are obtained when instead setting $\kappa_0=0.1, \kappa_0=10$ in a sensitivity analysis. Code is available at \url{https://github.com/stephenyang/CVB_NGFA}.
\subsection{Simulated data}

\begin{figure*}
\centering
\begin{tabular}{c} 
\begin{minipage}{0.6\textwidth}
 \includegraphics[width=10cm,height=6cm, keepaspectratio, angle=0]{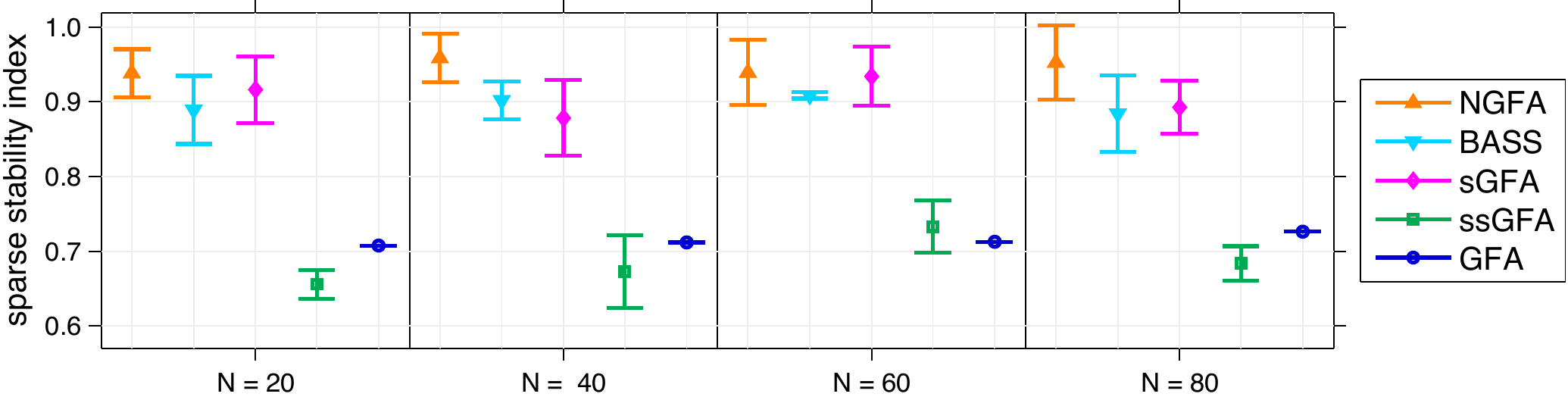}
\end{minipage}
\\
\begin{minipage}{0.6\textwidth}
 \includegraphics[width=10cm,height=6cm, keepaspectratio, angle=0]{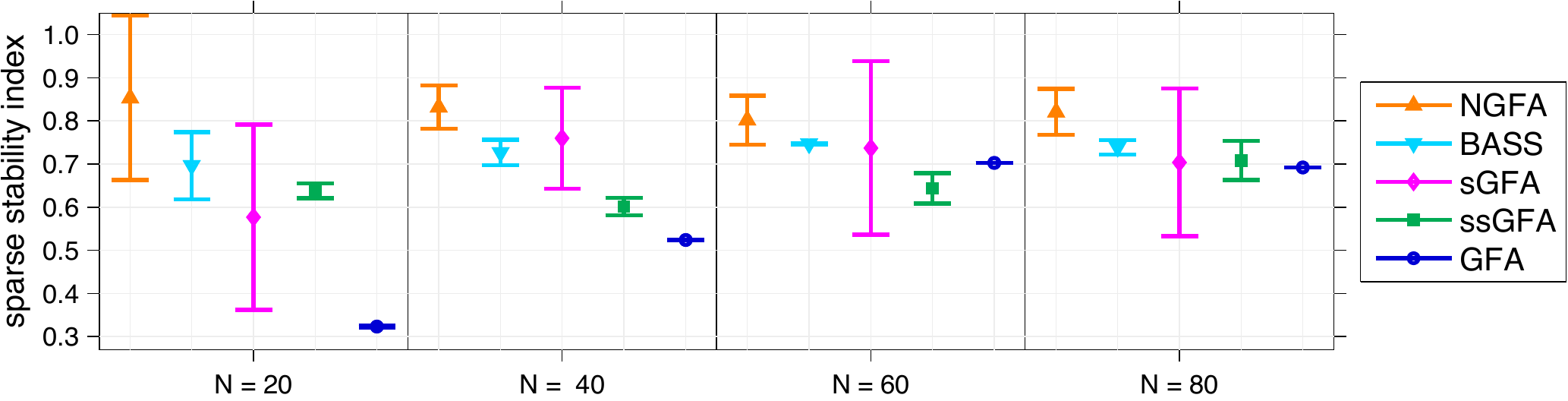}
\end{minipage}
\\
\begin{minipage}{0.6\textwidth}
 \includegraphics[width=10cm,height=6cm, keepaspectratio, angle=0]{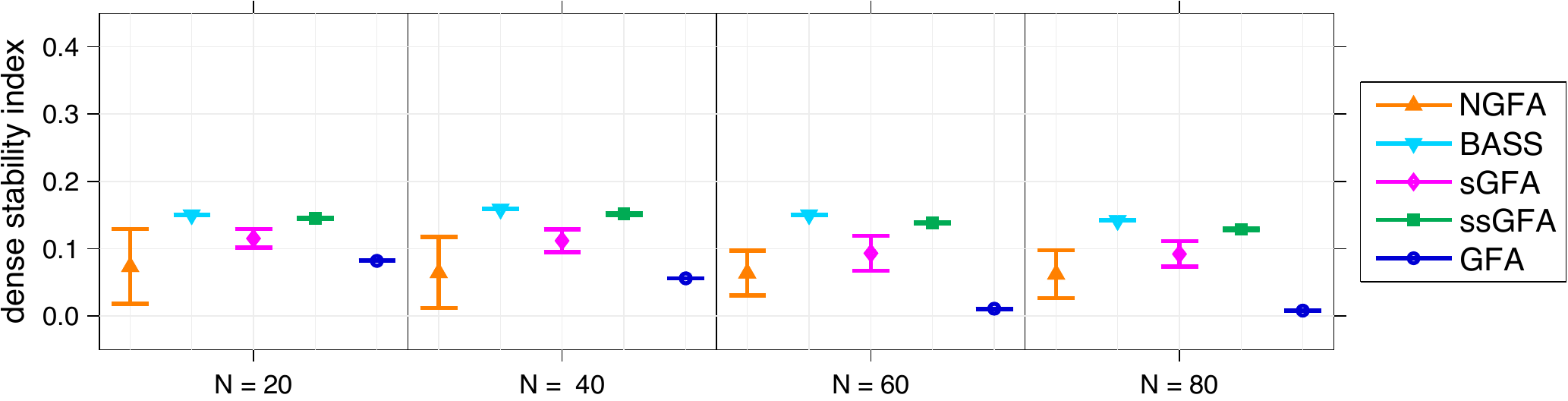}
\end{minipage}
\end{tabular}
\vspace{-0.5em}
\caption{The comparison of stability indices on the inferred matrix of factor loadings for our synthetic data. 
For {SSI}, higher is better; for {DSI}, lower is better. The means and the standard derivations of the stability indices are denoted by the marker and the bar respectively. The SSI comparison of all methods in Simulation 1 is shown in upper rows; The SSI and DSI comparisons in Simulation 2 are shown in middle and bottom rows, respectively.}
\label{SI}
\end{figure*}

\begin{figure*}
  \centering
   \begin{tabular}{l}
   \vspace{0em}
   \textbf{A}\\
    \begin{minipage}{0.75\textwidth}
      \includegraphics[width=20cm,height=10cm, keepaspectratio]{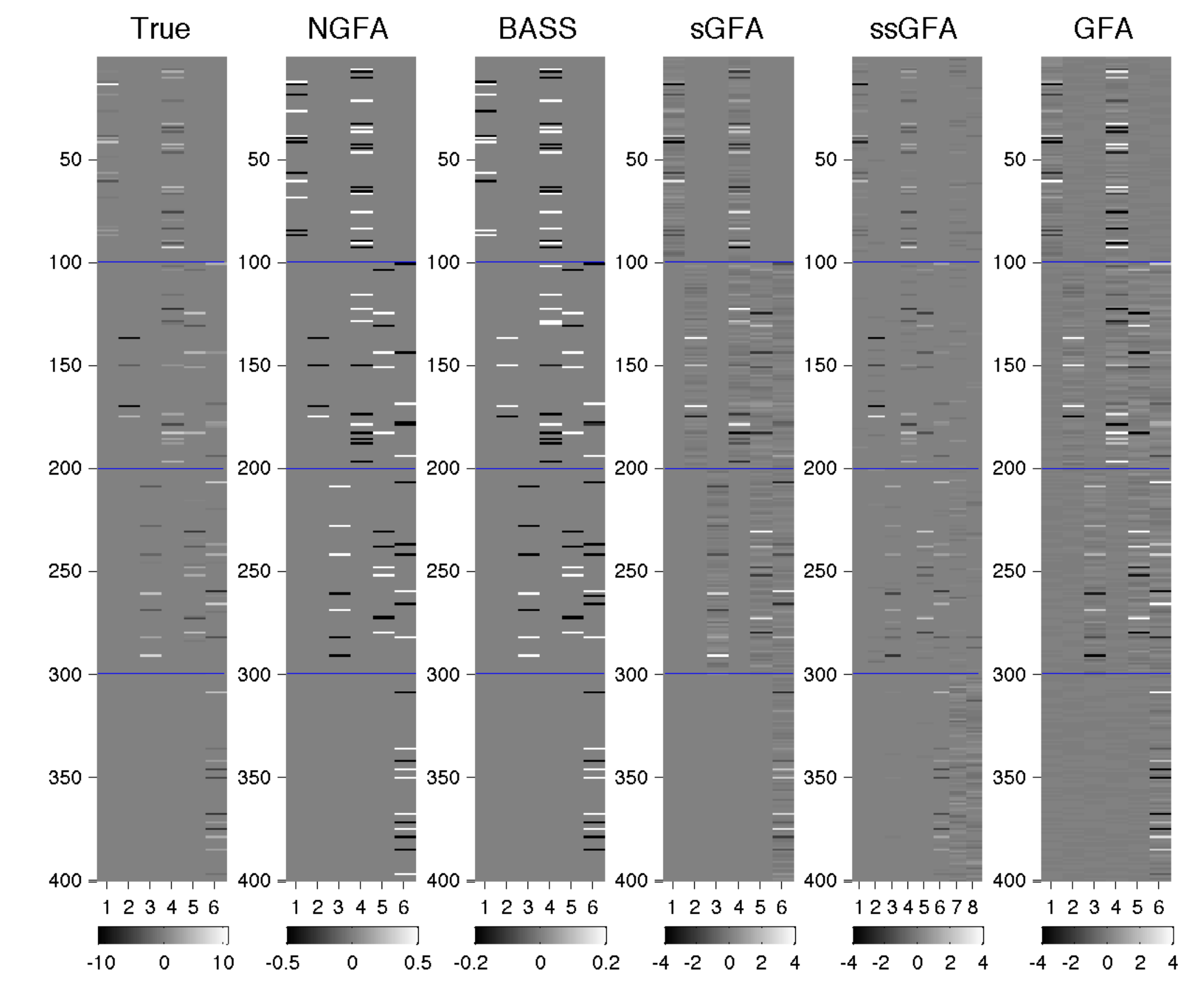}
    \end{minipage}
    \vspace{0em}
    \\
      \textbf{B}\\
      \begin{minipage}{0.75\textwidth}
       \includegraphics[width=20cm,height=10cm, keepaspectratio]{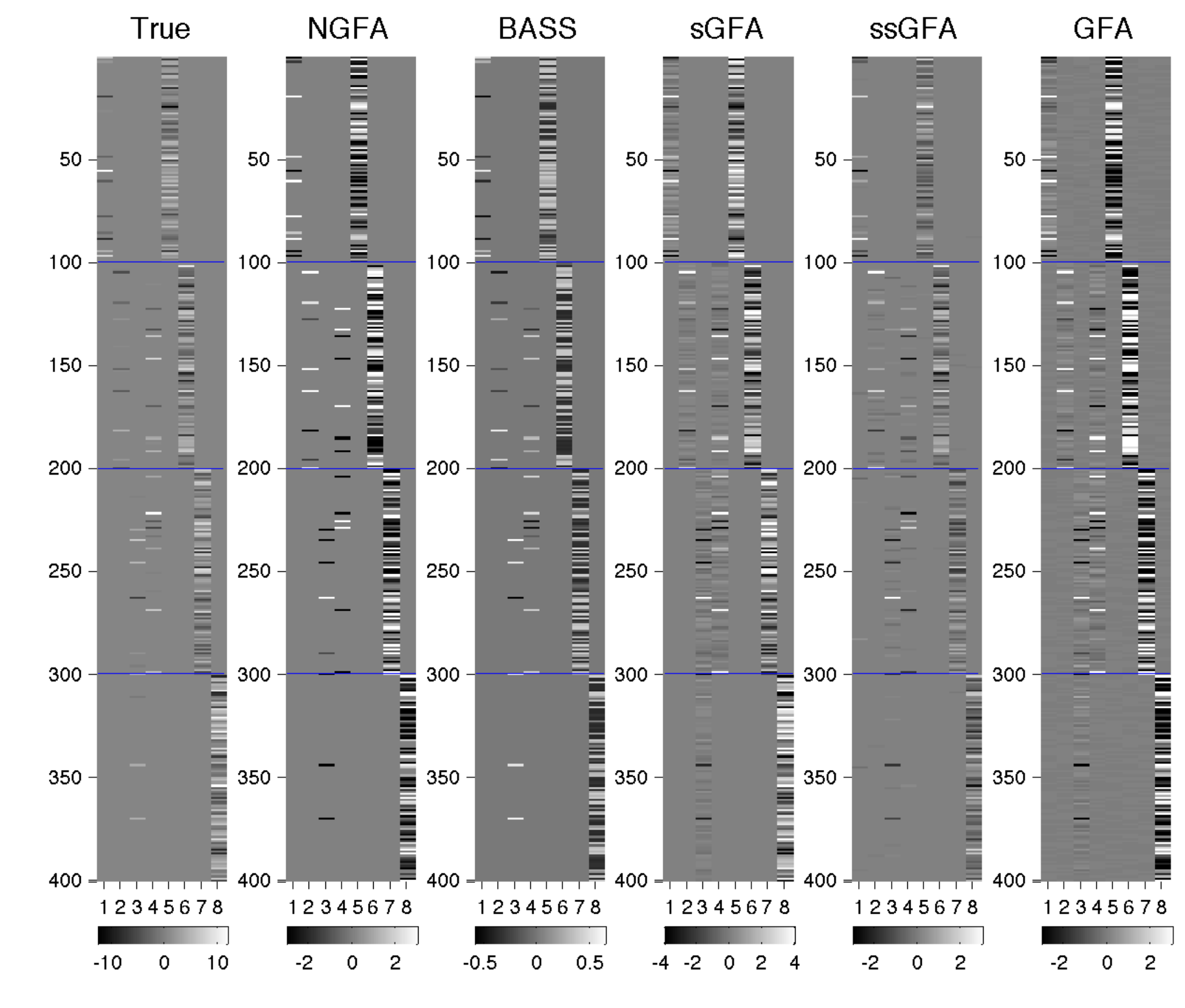}
    \end{minipage}
  \end{tabular}
\caption{The true and the inferred factor loadings by all methods in \emph{Simulation 1} (A) and \emph{Simulation 2} (B). The columns of the inferred factor loading matrices were reordered for easy comparison. The horizontal lines separate the four groups.}
\label{G_comp_01}
\end{figure*}

For our evaluations on synthetic data, we adopt the simulation study in \citep{BASS2016}: we perform two simulations (\emph{Simulation 1} and \emph{Simulation 2}) which include four groups of data with the dimensionality $D_{\scriptscriptstyle m} = 100$ for each group, respectively. The numbers of samples in the four groups are set to $N = \{20, 40, 60, 100\}$, respectively. In \emph{Simulation 1}, we set the number of latent factors $K = 6$, and generate data only with sparse factor loadings. Specifically, the first three factors are specific to $\mathbf{X}^{\scalebox{0.65}{(1)}}, \mathbf{X}^{\scalebox{0.65}{(2)}}$ and $\mathbf{X}^{\scalebox{0.65}{(3)}}$, respectively, and the last three are shared among all groups. In \emph{Simulation 2}, we set $K = 8$ and generate data with both sparse and dense factor loadings. The sparsity pattern is described in Table~\ref{sp_table}, and also shown in Fig.~\ref{G_comp_01}. 

\begin{table}[htp!]
	\centering	
	\small
	\begin{tabular}{ |m{0.32cm} | m{0.10535cm}  m{0.10535cm}  m{0.10535cm}  m{0.10535cm}  m{0.10535cm}  m{0.10535cm} | m{0.10535cm} m{0.10535cm}  m{0.10535cm} m{0.10535cm}  m{0.10535cm}  m{0.10535cm}  m{0.10535cm}  m{0.10535cm} |}
	\hline
	& \multicolumn{6}{c|}{Simulation 1} & \multicolumn{8}{c|}{Simulation 2} \\
	\hline 
  &1&2&3&4&5&6&1&2&3&4&5&6&7&8\\
\hline
$X^{\scalebox{0.625}{(1)}}$&s&-&-&s&-&-&s&-&-&-&d&-&-&-\\
$X^{\scalebox{0.625}{(2)}}$&-&s&-&s&s&s&-&s&-&s&-&d&-&-\\
$X^{\scalebox{0.625}{(3)}}$&-&-&s&-&s&s&-&-&s&s&-&-&d&-\\
$X^{\scalebox{0.625}{(4)}}$&-&-&-&-&-&s&-&-&s&-&-&-&-&d\\
\hline 	
\end{tabular}
\caption{\label{sp_table} {Sparsity pattern of the factor loading matrices in Simulation 1 and 2.} \enquote{s} represents a sparse column vector; \enquote{d} represents a dense column vector; \enquote{-} represents no contribution to that group from the factor.}
\end{table}

The sparsity of the sparse factor loadings is handled by setting $90\%$ of the entries in each loading column to zero at random, and the nonzero entries in both the sparse and dense factor loadings are generated from a Gaussian distribution $\mathcal{N}(0, 4)$. The latent factors are generated from a standard Gaussian distribution (i.e., zero mean and unit variance). We generate the residual noise i.i.d. from a Gaussian distribution $\mathcal{N}(0, 1)$. 

We compare the following methods: 
\textbf{(1) GFA:} The Bayesian group factor analysis model~\citep{BGFA2012} with column-wise ARD priors to induce column-wise sparsity on the factor loading matrix. For the GFA model, we used the \texttt{GFA} package with the default parameters setting as set in the code released online.~\footnote{https://cran.r-project.org/web/packages/GFA/index.html.} The initial number of factors is set to the true values. The optimization method is L-BFGS with the maximum iterations set to $10^5$.
\textbf{(2) sGFA:} The extension of the GFA with element-wise ARD priors inducing element-wise sparsity~\citep{SGFA2016}. For the sGFA model, the initial number of factors is set to half of the minimum of the sample size and the total number of variables, i.e., $K = \min(N, \sum_m{D}_m)$. The total number of {MCMC} iterations is set to $10^5$ with sampling steps set to $10^3$ and thinning steps set to $5$.
\textbf{(3) ssGFA:} The extension of the GFA with the spike-and-slab prior~\citep{SGFA2016}, for which we again use the \texttt{GFA} package with the spike-and-slab prior. We set the noise parameters by the \texttt{informativeNoisePrior} function to prevent overfitting. The initial number of factors is set to half of the minimum of the sample size and the total number of variables. The total number of MCMC iterations is set to $10^5$ with sampling steps set to $10^3$ and thinning steps set to $5$.
\textbf{(4) BASS:} The Bayesian group factor analysis with structured sparsity priors (BASS)~\citep{BASS2016}, for which we use the code released in~\citep{BASS2016}.~\footnote{https://github.com/judyboon/BASS.} The BASS is initialized using 50 iterations of MCMC and followed by expectation maximization until convergence, reached when both the number of nonzero loadings do not change for $t$ iterations and the log-likelihood change is less than $1\times10^{-5}$ within $t$ iterations. The initial number of factors is set to 10 in \emph{Simulation} 1 and 15 in \emph{Simulation} 2 as described in \citep{BASS2016}. 
%
We perform 20 runs for each method, in particular to evaluate the sensitivity of our inference algorithm to initialization since CVI algorithms are only guaranteed to converge to a local optimum. For all the experiments, we simply set the initial number of factors for our method to be the minimum of the sample size and the dimensionality of each group, and run the model with CVI algorithm until convergence.
%

To evaluate the performance of the methods on the recovery of sparse and dense factor loadings, we use the sparse and dense stability index defined in \citep{BASS2016} to quantify the distance between the true and the inferred factor loading matrices. Given the absolute correlation matrix $\mathbf{C} \in \mathbb{R}^{K_1 \times K_2}$ of the columns of two sparse matrices, the \emph{sparse stability index} (SSI) is calculated as
\begin{align}
\mathrm{SSI} 
= &\ \frac{1}{2K_1} \sum_{\scriptscriptstyle r=1}^{K_1} \Big( \max(\mathbf{C}_{\scriptscriptstyle r \scalebox{0.75}{:} }) - \frac{\sum_{\scriptscriptstyle l} \mathbbm{1}(\mathbf{C}_{\scriptscriptstyle rl} > \hat{\mathbf{C}}_{\scriptscriptstyle r\scalebox{0.75}{:}})\mathbf{C}_{\scriptscriptstyle rl}}{K_2-1} \Big) \notag\\
+&\ \frac{1}{2K_2} \sum_{\scriptscriptstyle l=1}^{K_2} \Big( \max(\mathbf{C}_{\scriptscriptstyle \scalebox{0.75}{:}l}) - \frac{\sum_{r} \mathbbm{1}(\mathbf{C}_{\scriptscriptstyle rl} > \hat{\mathbf{C}}_{\scriptscriptstyle \scalebox{0.75}{:}l})\mathbf{C}_{\scriptscriptstyle rl}}{K_1-1} \Big),\notag
\end{align}
where $\mathbf{C}_{\scriptscriptstyle r\scalebox{0.75}{:}}$ and $\mathbf{C}_{\scriptscriptstyle \scalebox{0.75}{:}l}$ denote the $r$-th row and $l$-th column of the matrix $\mathbf{C}$, respectively; $\hat{\mathbf{C}}_{\scriptscriptstyle r\scalebox{0.75}{:}}$ and $\hat{\mathbf{C}}_{\scriptscriptstyle \scalebox{0.75}{:}l}$ denote the mean of the $r$-th row and $l$-th column of the matrix $\mathbf{C}$, respectively.
The SSI is invariant to column-scaling and -permutation; larger values indicate better recovery.

The \emph{dense stability index} (DSI) measures the distance between dense matrix columns. Given two dense matrices $\mathbf{M}_1 \in \mathbb{R}^{K_1\times D}$ and $\mathbf{M}_2 \in \mathbb{R}^{K_2 \times D}$, the DSI is defined as
\begin{align}
\mathrm{DSI} &\ = \frac{1}{D^2} \mathrm{tr}(\mathbf{M}_1\mathbf{M}_1^T - \mathbf{M}_2 \mathbf{M}_2^T). \notag
\end{align}

The DSI is invariant to orthogonal matrix transformation, column-scaling and -permutation; the lower values indicate better recovery.

Following the strategy in~\citep{BASS2016}, in \emph{Simulation 1} where all factor loadings are sparse, we calculate the SSI between the true and recovered factor loading matrices. In \emph{Simulation 2}, we first threshold the recovered factor loading matrix entries with a sparsity threshold set to 0.15. Then, we categorize the columns of each recovered factor loading matrix into sparse columns and dense columns by selecting the first 4 columns with most nonzero entries as dense columns, and the remaining columns as sparse columns. We calculate SSI between the true and the recovered sparse factor loading columns, and DSI between the true and the recovered dense columns. We calculate the two stability indices for each group separately and average the result for all groups.

\begin{figure*} 
  \centering
   \begin{tabular}{  m{-1cm}  c c } 
  & {\scriptsize Hyman} & {\scriptsize Pollack}\\ 
    \vspace{-2.5em}
    \rotatebox[origin=Bc]{90}{\parbox{5cm}{{\scriptsize \ \ \ \ \ \ \ \ \ \ \ \ \ \ \ \ \ \ \ \ \ \ \ \ \  \ \ \ \ \ \ \ \ \ \ \ \ \ \  \ \ AUC}}} 
    &
    \begin{minipage}{.35\textwidth}
      \includegraphics[width=6cm,height=5cm, keepaspectratio, angle=90]{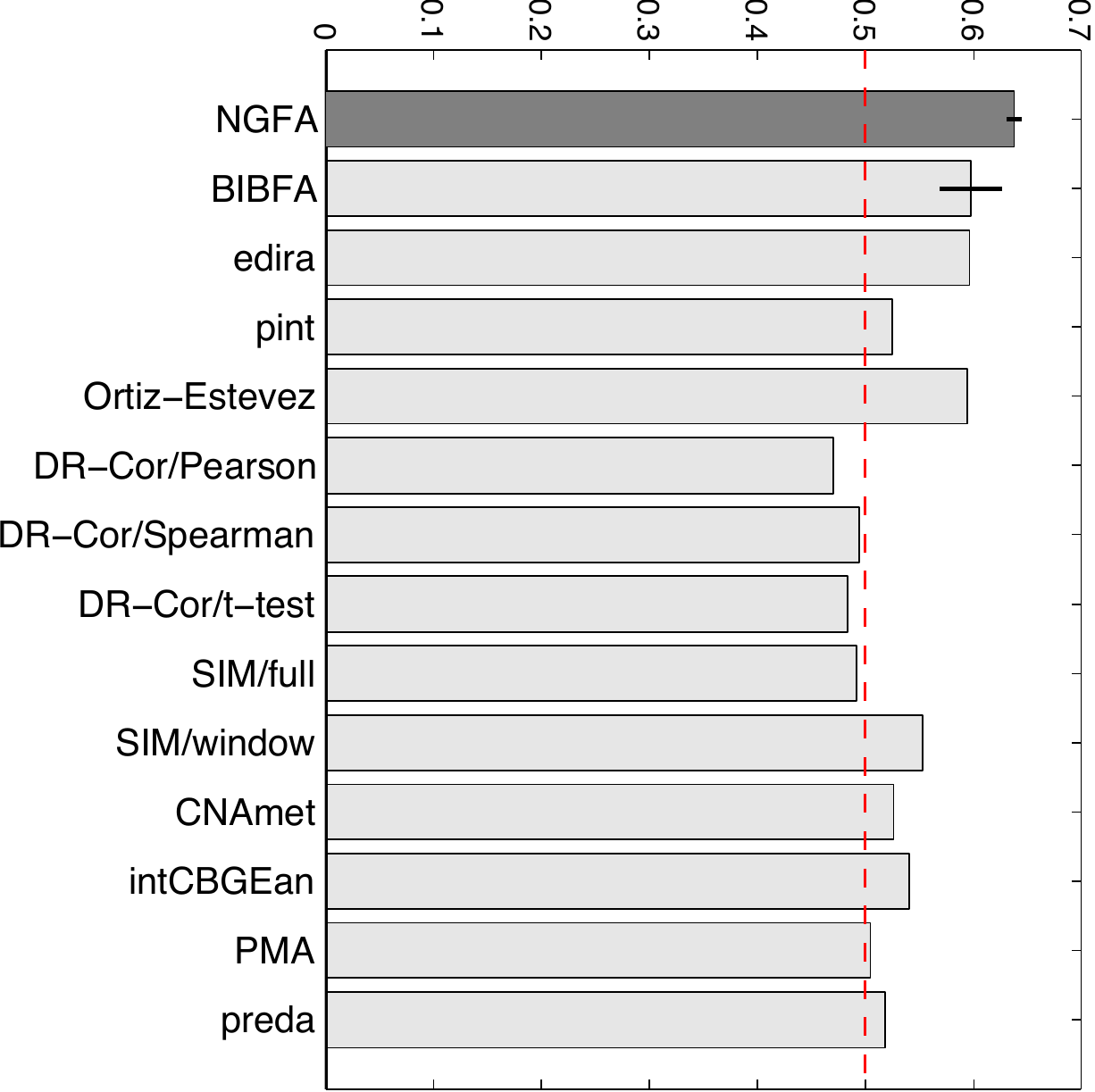}
    \end{minipage}
    &
      \begin{minipage}{.35\textwidth}
      \includegraphics[width=6cm,height=5cm, keepaspectratio, angle=90]{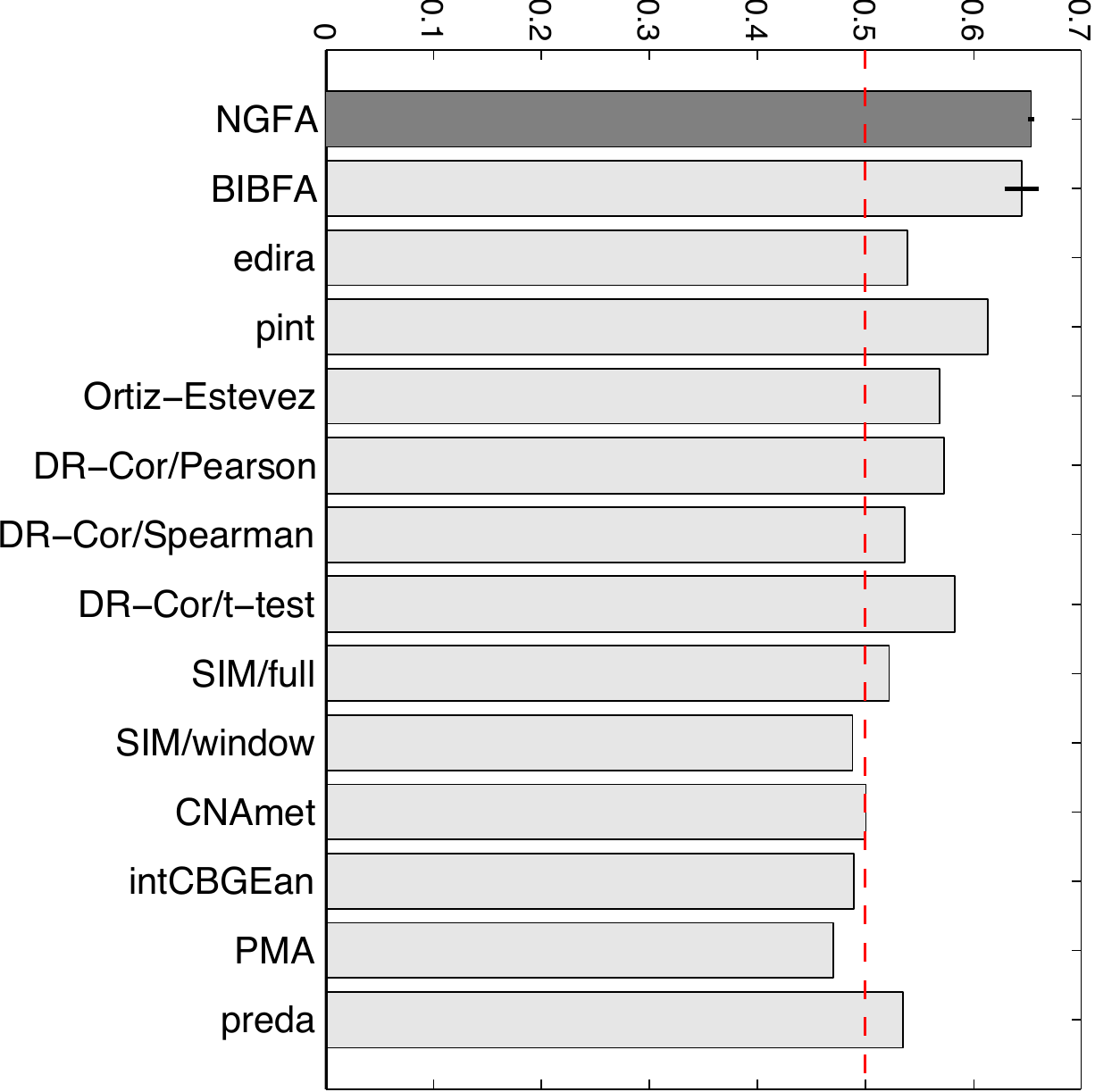}
    \end{minipage}
  \end{tabular}
  \vspace{-1em}
  \caption{Evaluation of cancer gene prioritization performance of the methods on two data sets: Hyman (left) and Pollack (right). The result is quantified by the area under the ROC curve (AUC). The dashed line indicates the {AUC} score for a random list ({AUC} = 0.5). The comparison shows that the NGFA achieves best performance.}
  \label{CGP}
\end{figure*}

The true and the inferred factor loading matrices by all methods in \emph{Simulation 1} and \emph{Simulation 2} are shown in Fig.~\ref{G_comp_01}.
The ARD prior cannot induce sufficient sparsity by pushing irrelevant factor loadings to small values. As a consequence, the GFA has difficulty in recovering sparse factor loadings because of the columns-wise ARD priors (Fig.~\ref{G_comp_01}). Similarly, the sGFA cannot induce sufficient element-wise sparsity within the loading columns by the independent ARD priors (Fig.~\ref{G_comp_01}). The ssGFA overfitted to data by not sufficiently shutting off the redundant factors (Fig.~\ref{G_comp_01}). Both the BASS and NGFA achieve element-wise sparsity effectively (Fig.~\ref{G_comp_01}). We quantify the performance of the methods with stability indices, i.e., the means and the standard derivations of the stability indices for each method over 20 runs are shown in Fig.~\ref{SI}. The NGFA using our CVI algorithm achieves the best SSI and DSI scores almost for all sample sizes. 
\subsection{Cancer gene prioritization}
Integrative analysis of multiple genomic data sets for understanding the genetic basis of common diseases has been challenging. For instance, DNA alterations that are frequent in cancers, measured by copy number variation (CNV) data, are known to induce gene expression modifications. Hence, cancer-related genes can be discovered by searching for such interactions. Recently, Bayesian GFA methods were applied to the task of cancer gene prioritization with encouraging results~\citep{BCCA2013}.
 To demonstrate the effectiveness of the NGFA using our CVI algorithm, we choose the same datasets \texttt{Hyman} and \texttt{Pollack} from~\citep{CGP2012} that are based on gene expression (GE) and CNV data as described in Table~\ref{cancergene_table}.
\begin{table}[H]
	\centering	
	\small	
	\begin{tabular}{ | p{1cm} | m{1.7cm} | p{1.7cm} | p{1.8cm} |}
	\hline
	Dataset  & \# genes & \# samples & \# cancer genes  \\
	\hline 
Hyman & 7489 & 14 & 48  \\ \hline
Pollack & 4287 & 41 & 38  \\
	    \hline
 	\end{tabular}
		\caption{\label{cancergene_table}The details of cancer genomics datasets.} 
\end{table}
More specifically, we consider the patients as co-occurring samples and all the genes in the whole genome as features. The GE and CNV data constitute the two groups. 
We then rank the genes according to the quantity defined by $s_d = \sum_{k=1}^{K}|\mathsf{E}({g}_{kd}^{\scriptscriptstyle(1)})\mathsf{E}({g}_{kd}^{\scriptscriptstyle(2)})|$, that is the correlation between GE and CNV data captured by the shared factors.
We repeat the data pre-processing procedure in \citep{CGP2012}, and evaluate the model performance by the area under the curve of the receiver operating characteristic (AUC) for retrieving known cancer-related genes. We run the NGFA 20 times with the initial $K$ set to the minimum of the sample size and feature dimension. We compare the NGFA using CVI algorithm to the Bayesian inter-battery factor analysis (BIBFA) model. We run the BIBFA for 20 times according to the setting described in~\citep{BCCA2013}. The mean AUC scores and the standard deviations are shown in Fig.~\ref{CGP}. The AUC scores for all the other methods are cited from \citep{CGP2012} where the standard deviations cannot be presented because those alternatives are deterministic methods. The NGFA using our CVI algorithm outperforms all the alternative methods.
\subsection{Decoding fMRI brain activity}
\begin{figure} [htb]
  \centering
  \begin{tabular}{  m{1cm}  c } 
    \rotatebox[origin=Bc]{0}{\parbox{2cm}{{Presented}}} 
    &
    \begin{minipage}{0.035\textwidth}
\includegraphics[width=0.95\textwidth]{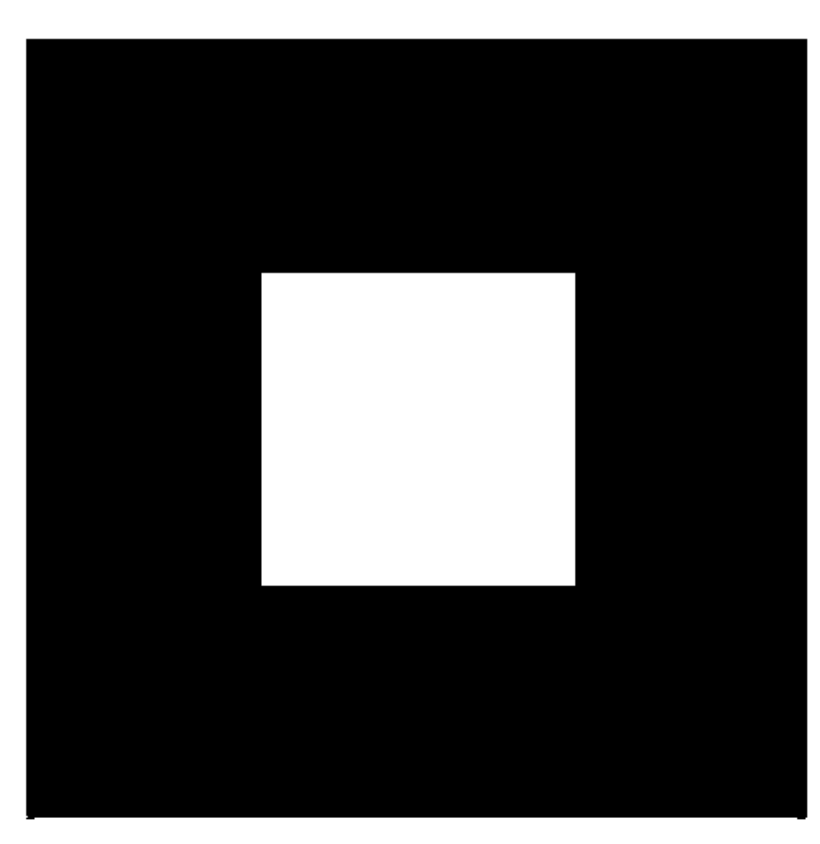}
\end{minipage}\hspace{-0.2em}
\begin{minipage}{0.035\textwidth}
\includegraphics[width=0.95\textwidth]{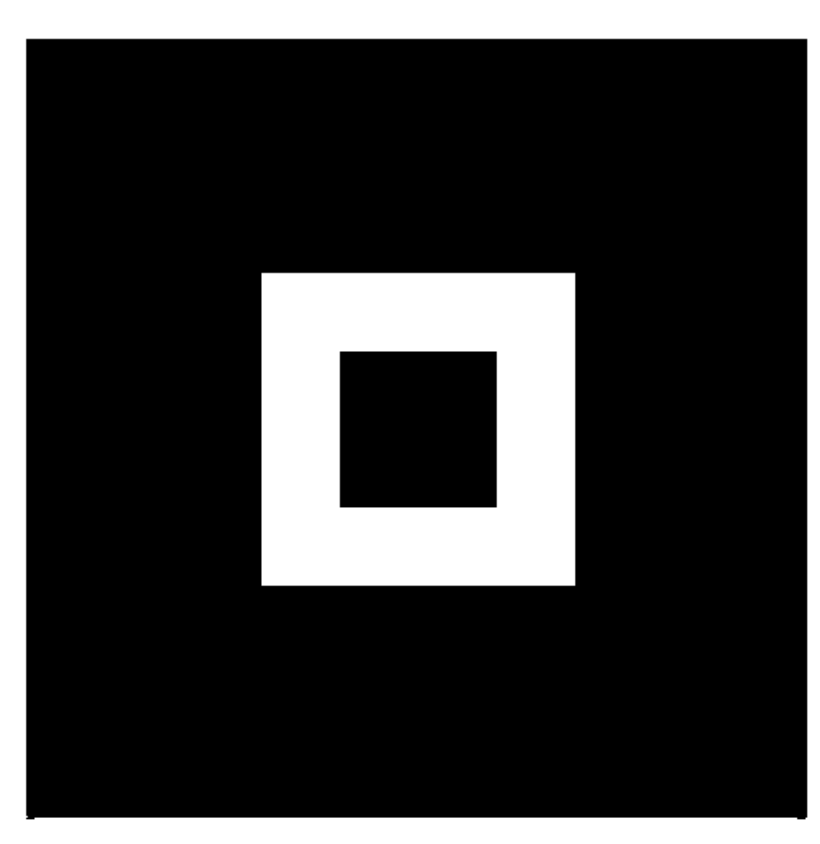}
\end{minipage}\hspace{-0.2em}
\begin{minipage}{0.035\textwidth}
\includegraphics[width=0.95\textwidth]{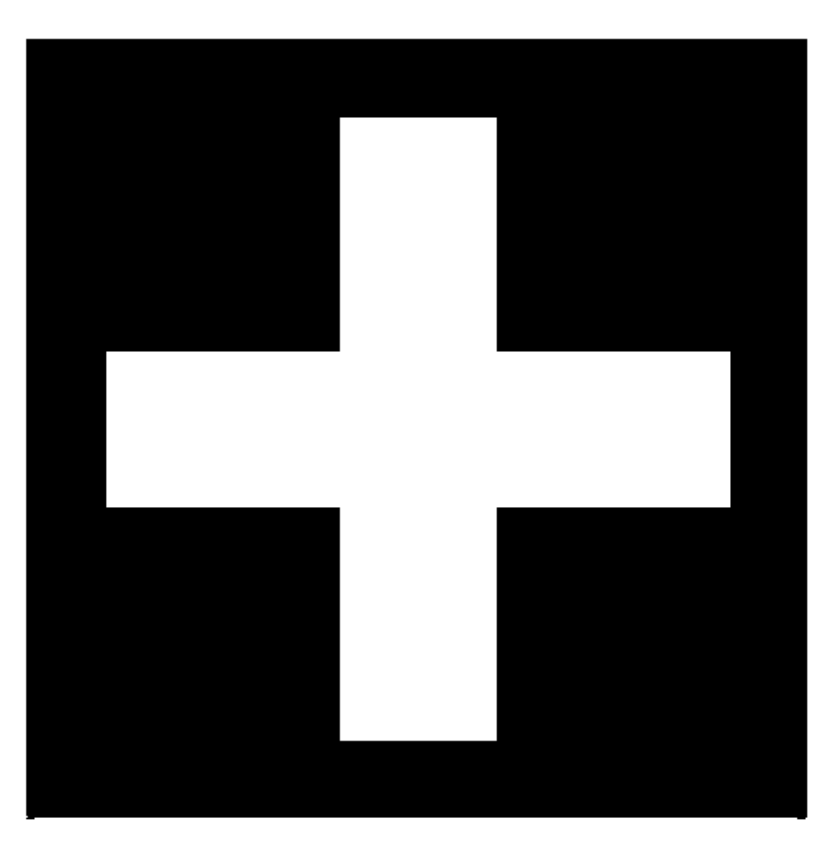}
\end{minipage}\hspace{-0.2em}
\begin{minipage}{0.035\textwidth}
\includegraphics[width=0.95\textwidth]{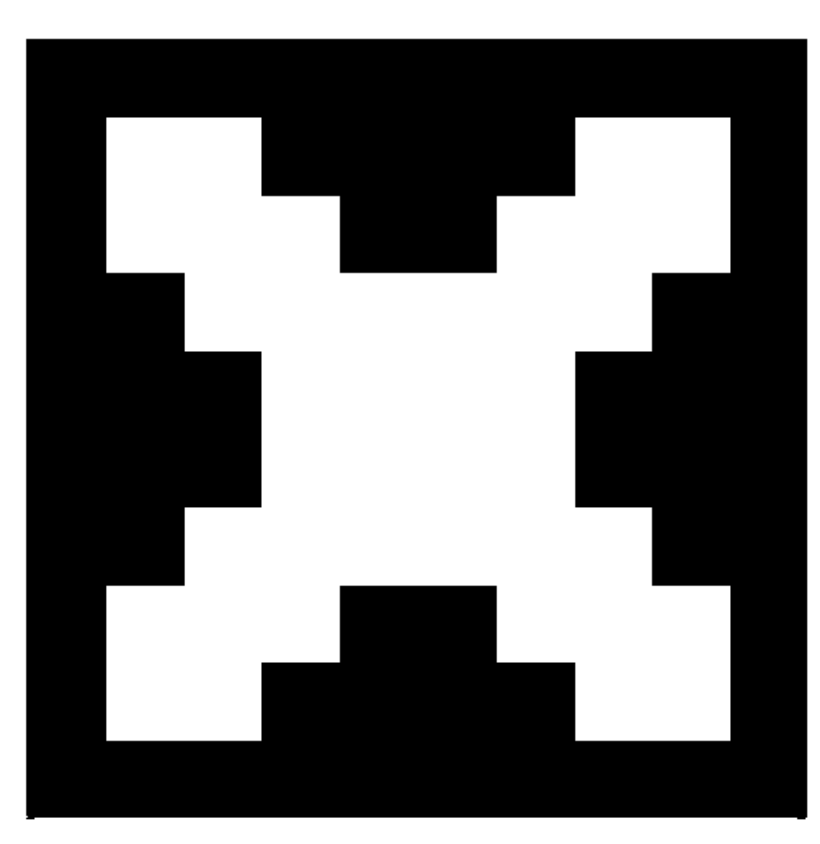}
\end{minipage}\hspace{-0.2em}
\begin{minipage}{0.035\textwidth}
\includegraphics[width=0.95\textwidth]{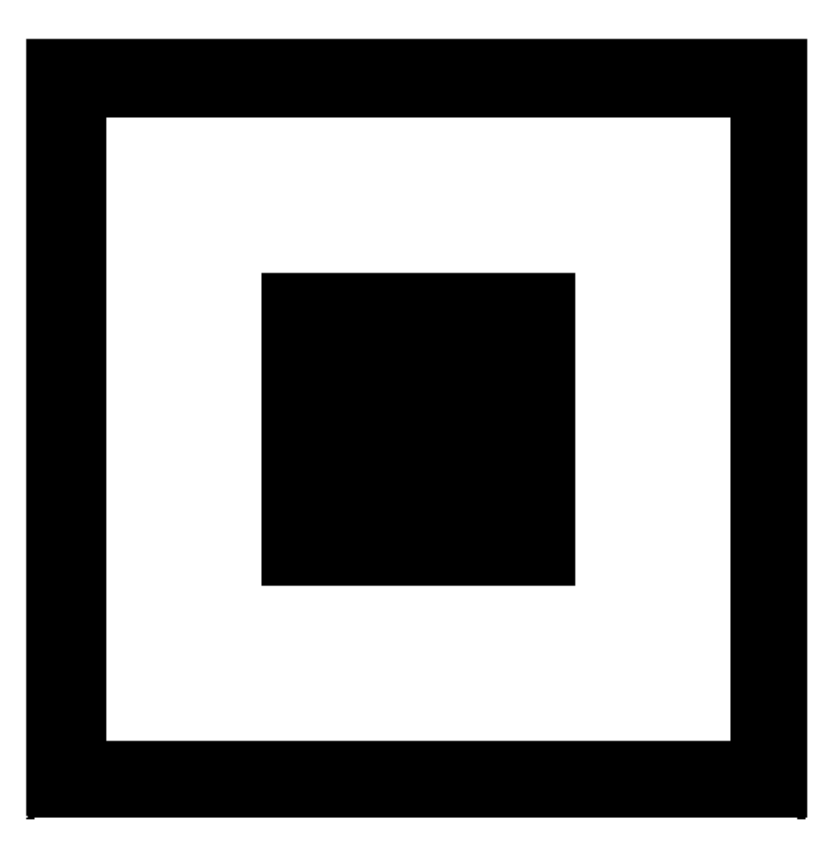}
\end{minipage}\hspace{-0.2em}
\begin{minipage}{0.035\textwidth}
\includegraphics[width=0.95\textwidth]{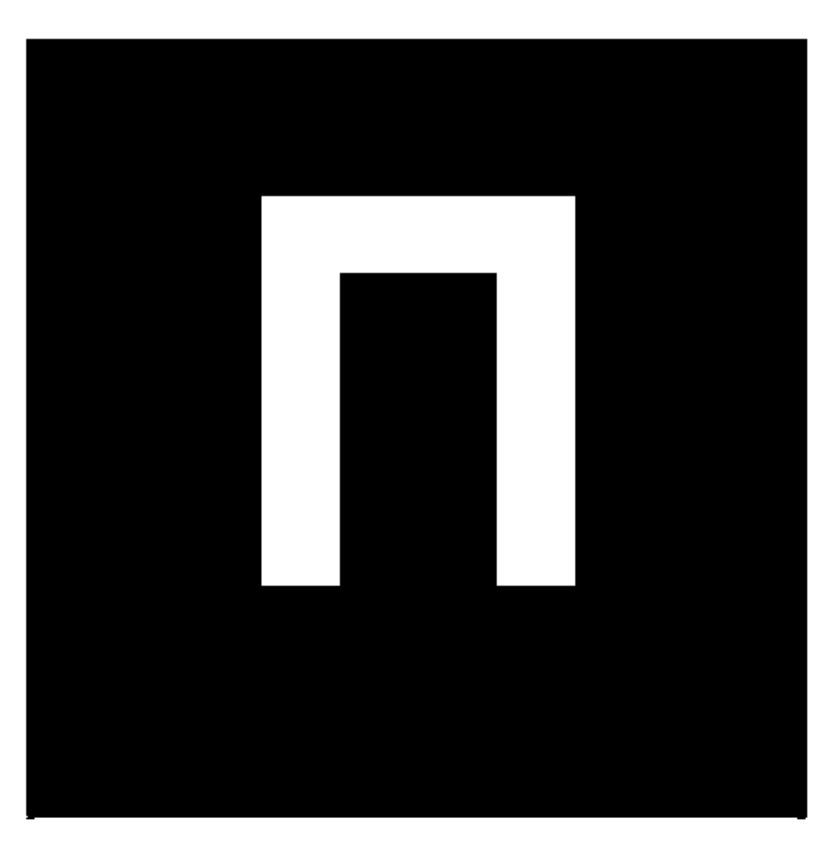}
\end{minipage}\hspace{-0.2em}
\begin{minipage}{0.035\textwidth}
\includegraphics[width=0.95\textwidth]{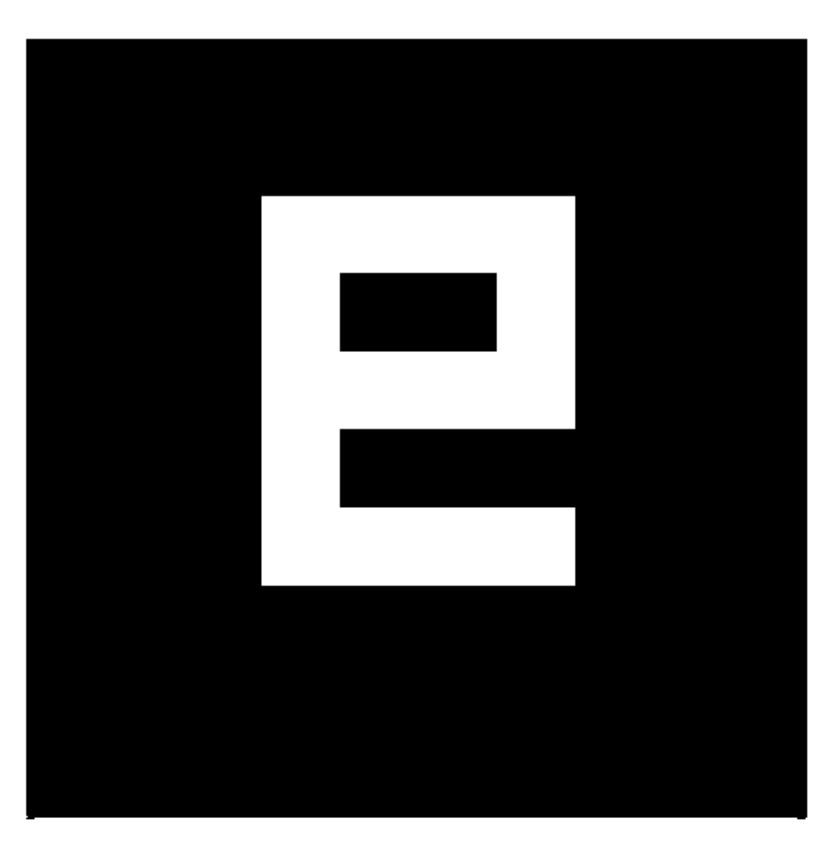}
\end{minipage}\hspace{-0.2em}
\begin{minipage}{0.035\textwidth}
\includegraphics[width=0.95\textwidth]{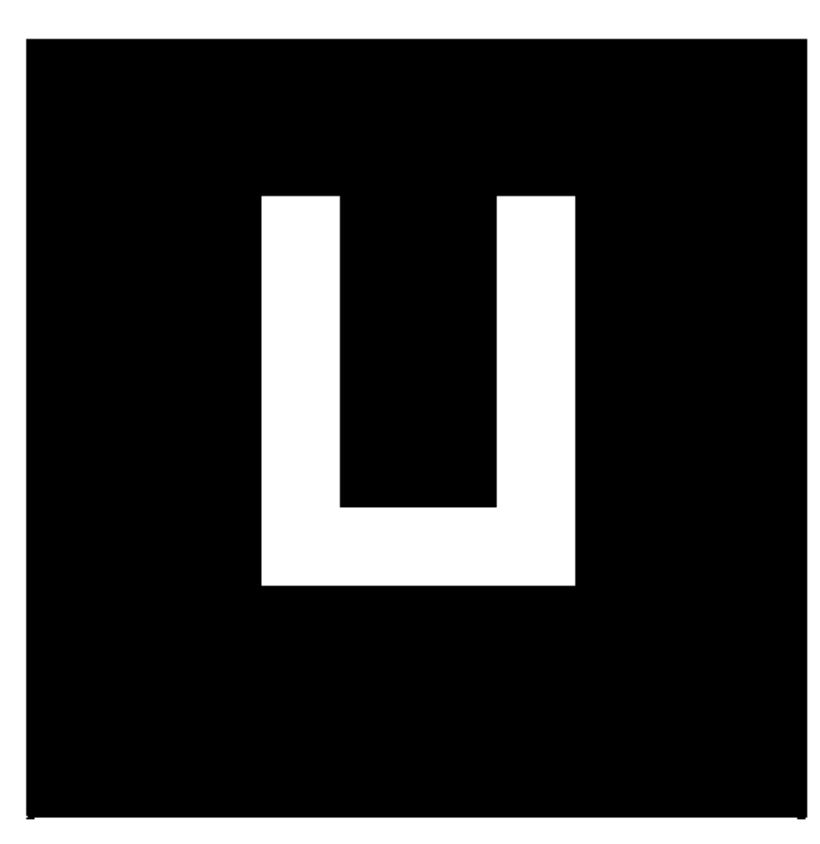}
\end{minipage}\hspace{-0.2em}
\begin{minipage}{0.035\textwidth}
\includegraphics[width=0.95\textwidth]{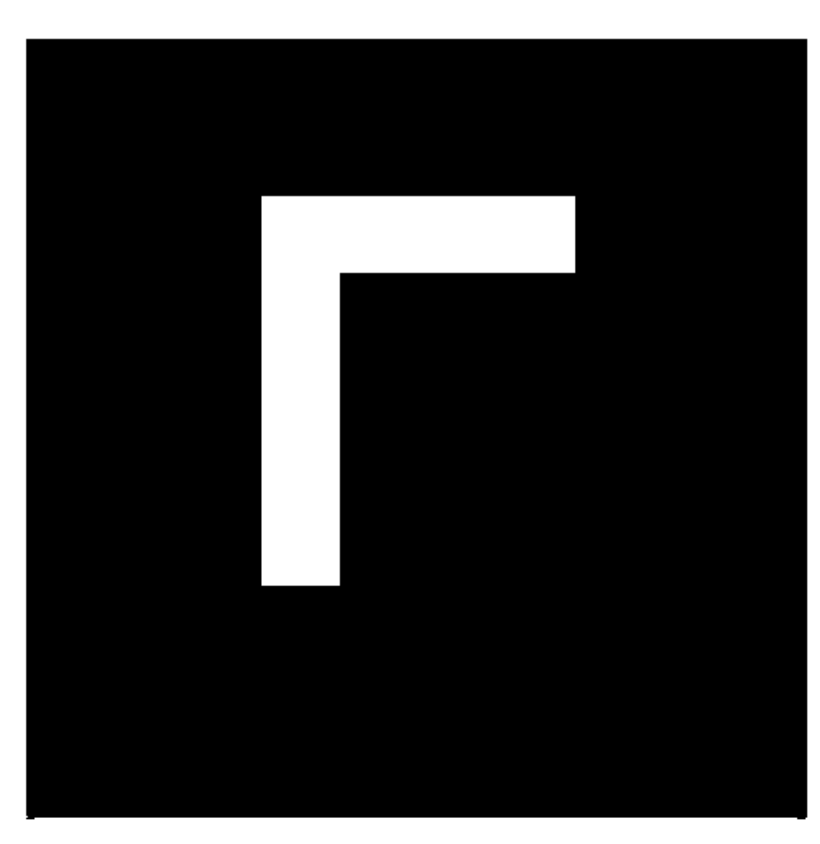}
\end{minipage}\hspace{-0.2em}
\begin{minipage}{0.035\textwidth}
\includegraphics[width=0.95\textwidth]{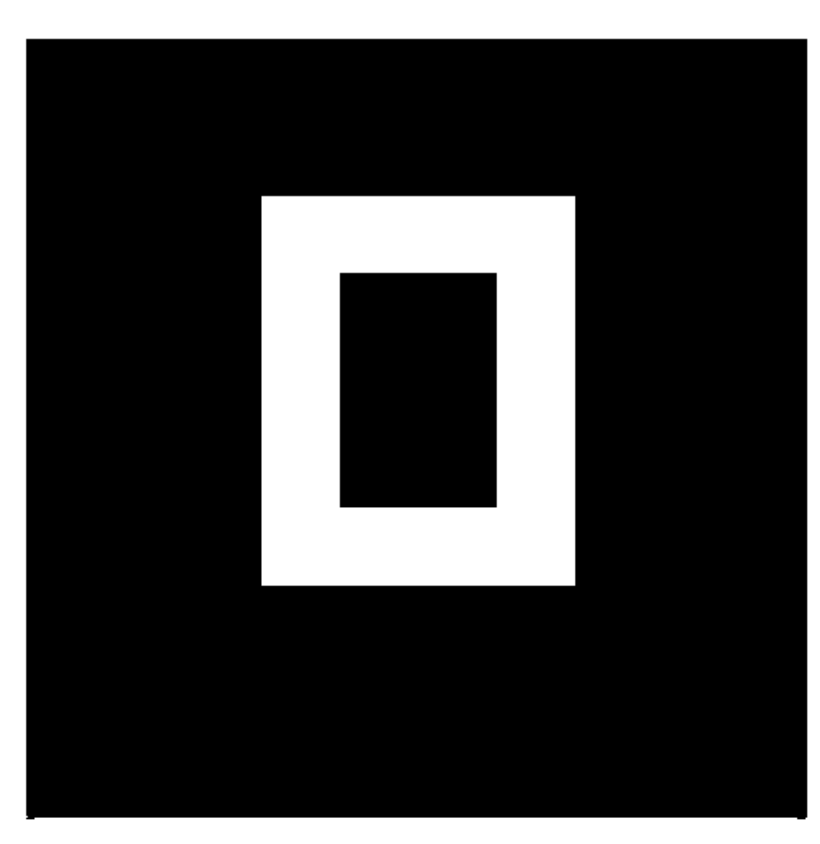}
\end{minipage}\hspace{-0.2em}
    \\
    \rotatebox[origin=Bc]{0}{\parbox{2cm}{{BCCA}}}
    &
\begin{minipage}{0.035\textwidth}
\includegraphics[width=0.95\textwidth]{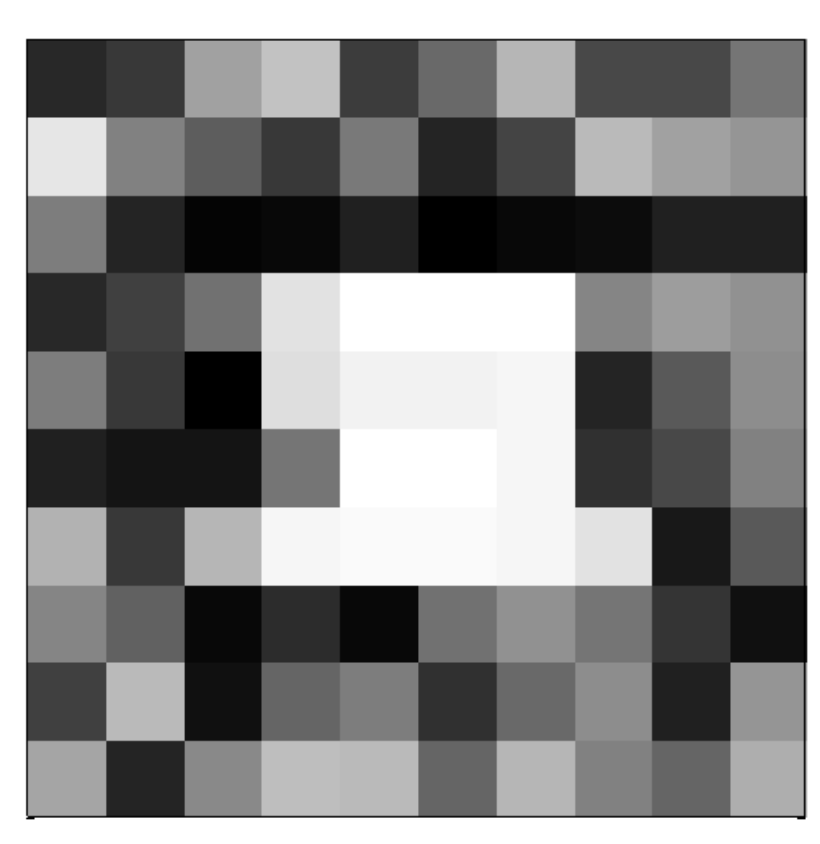}
\end{minipage}\hspace{-0.2em}
\begin{minipage}{0.035\textwidth}
\includegraphics[width=0.95\textwidth]{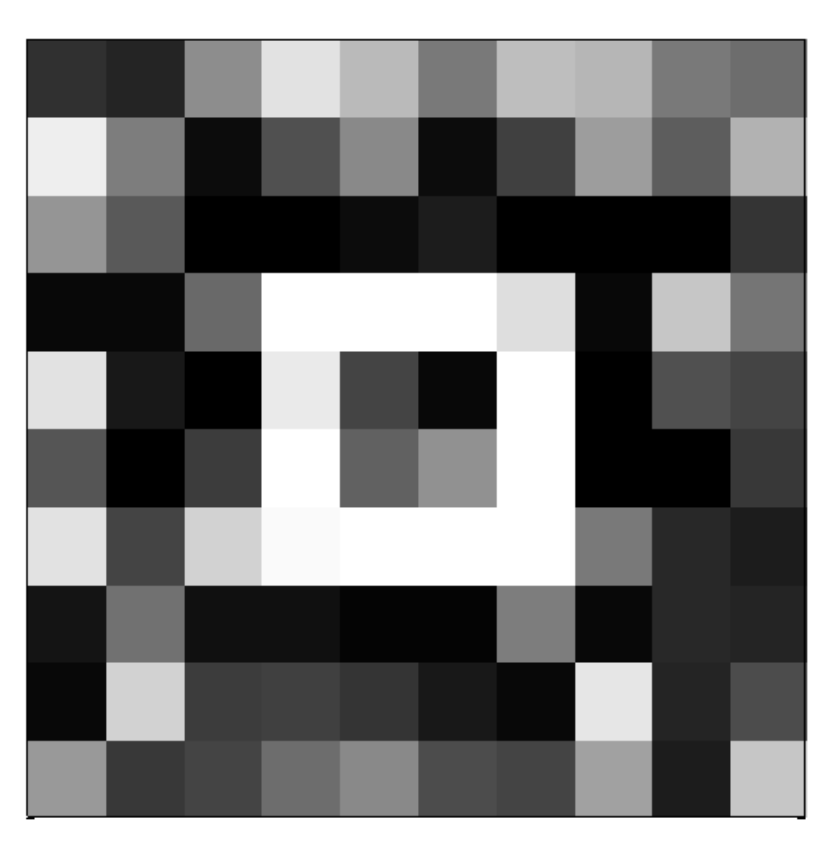}
\end{minipage}\hspace{-0.2em}
\begin{minipage}{0.035\textwidth}
\includegraphics[width=0.95\textwidth]{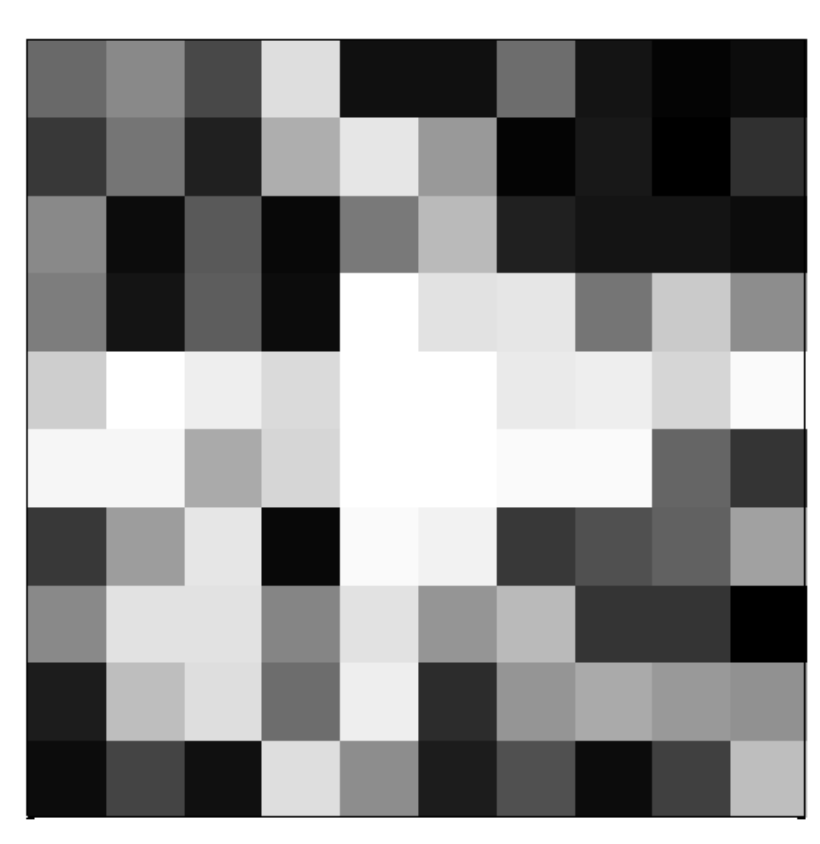}
\end{minipage}\hspace{-0.2em}
\begin{minipage}{0.035\textwidth}
\includegraphics[width=0.95\textwidth]{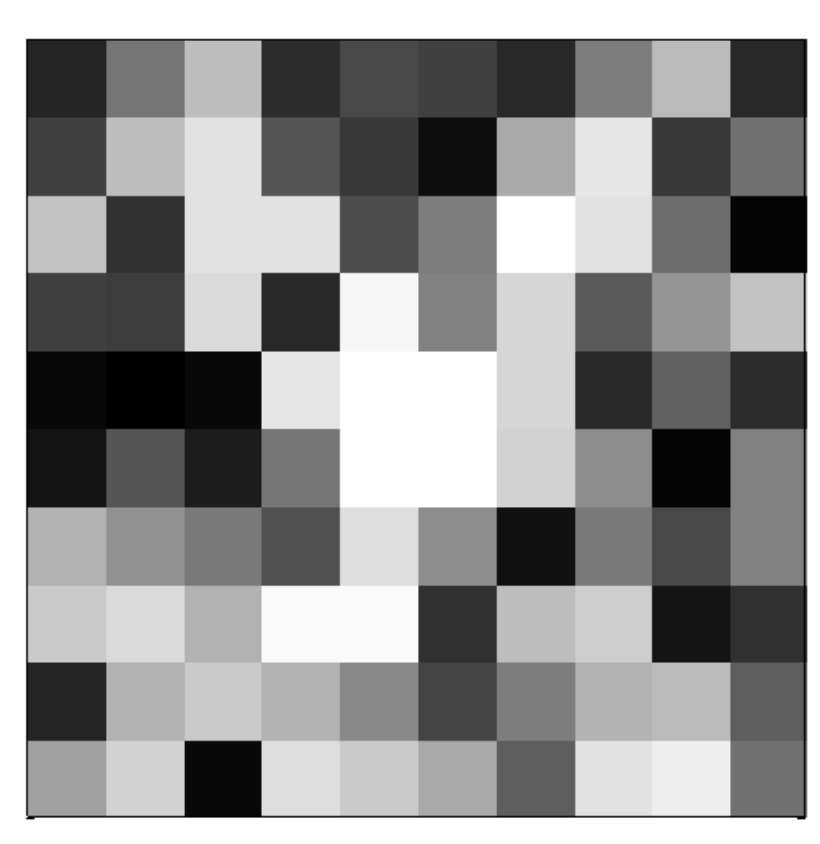}
\end{minipage}\hspace{-0.2em}
\begin{minipage}{0.035\textwidth}
\includegraphics[width=0.95\textwidth]{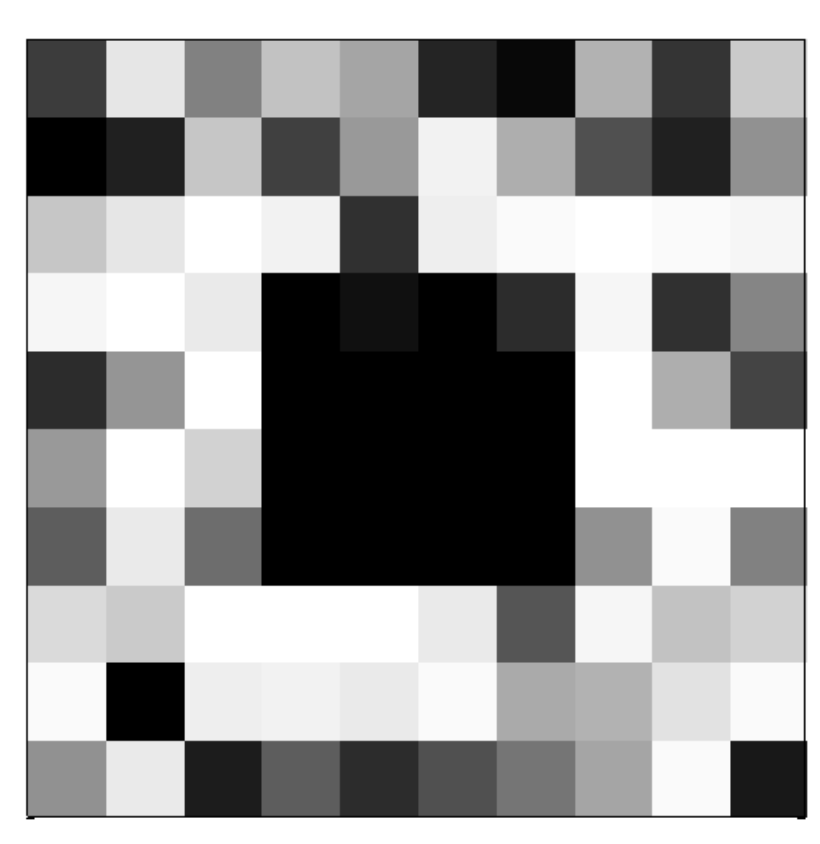}
\end{minipage}\hspace{-0.2em}
\begin{minipage}{0.035\textwidth}
\includegraphics[width=0.95\textwidth]{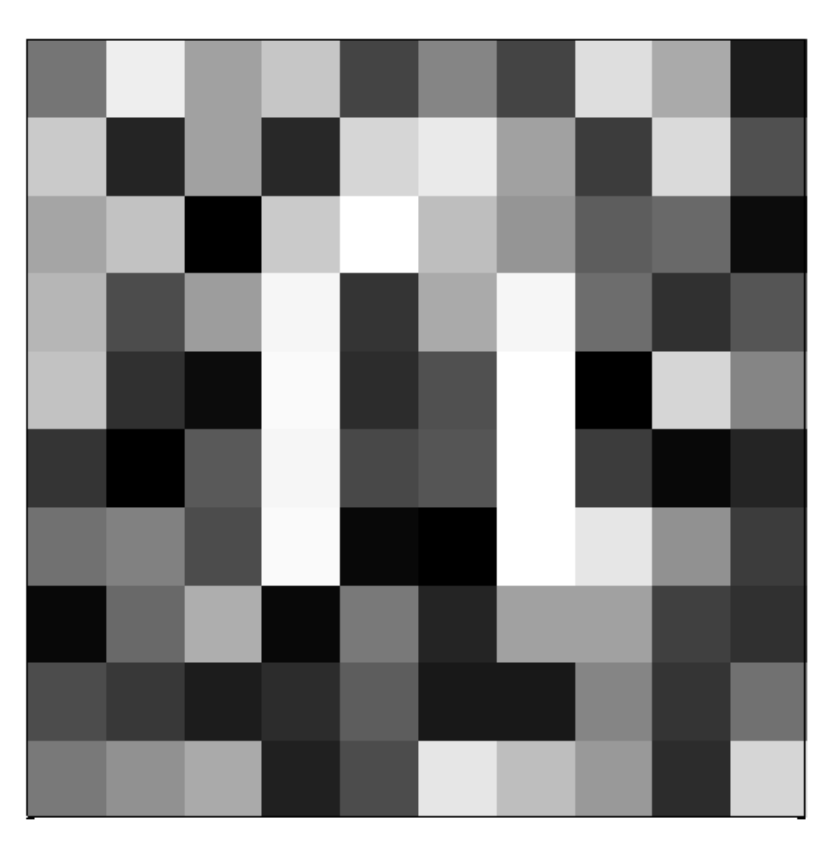}
\end{minipage}\hspace{-0.2em}
\begin{minipage}{0.035\textwidth}
\includegraphics[width=0.95\textwidth]{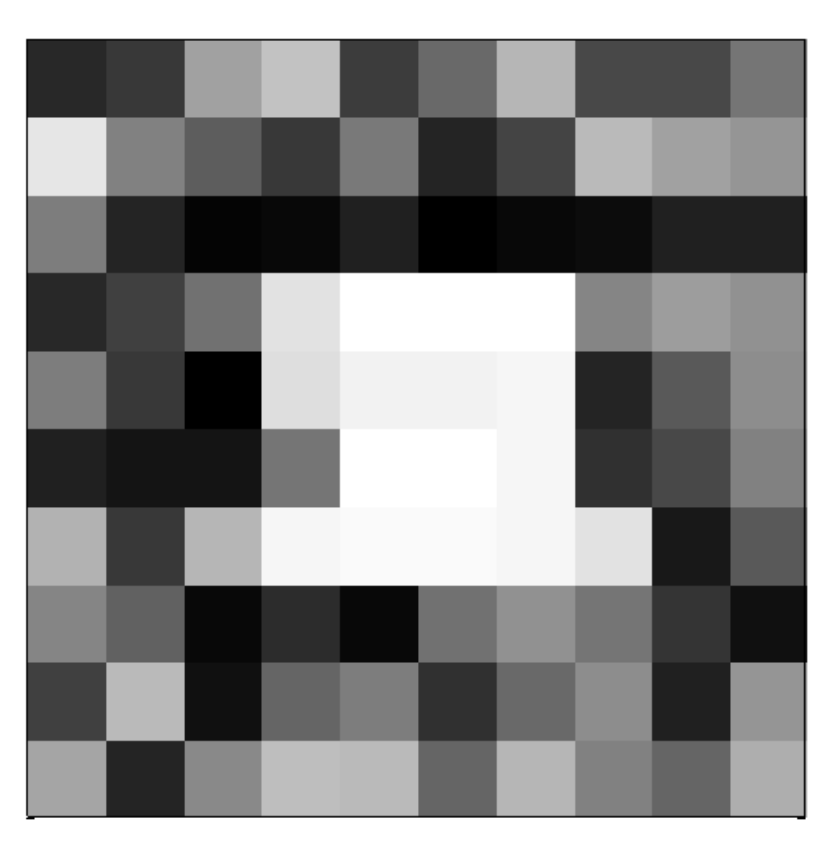}
\end{minipage}\hspace{-0.2em}
\begin{minipage}{0.035\textwidth}
\includegraphics[width=0.95\textwidth]{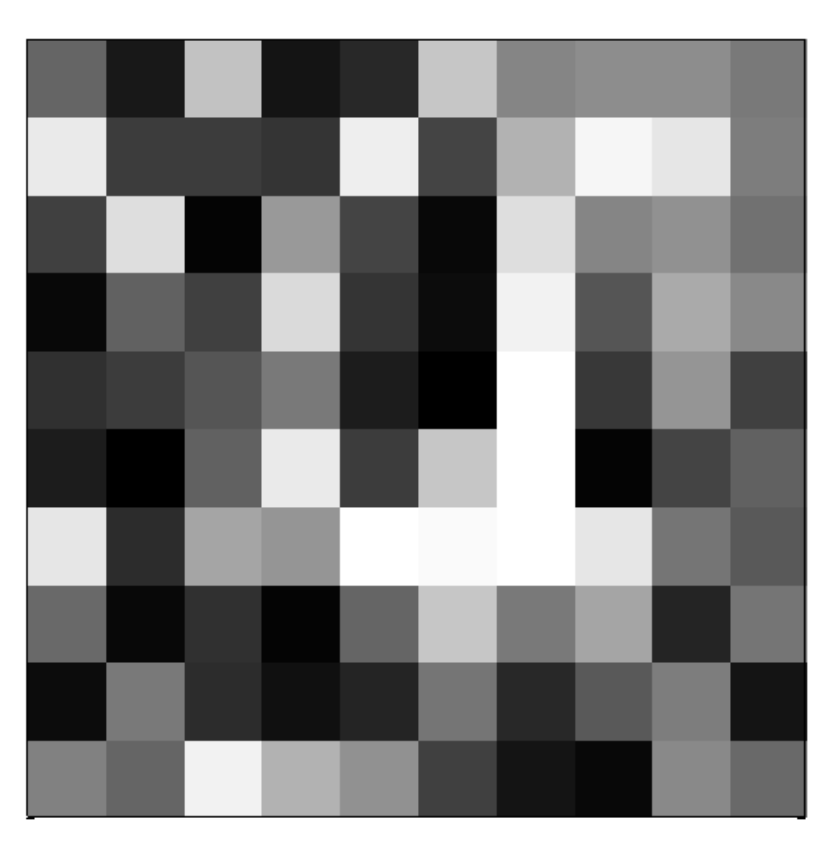}
\end{minipage}\hspace{-0.2em}
\begin{minipage}{0.035\textwidth}
\includegraphics[width=0.95\textwidth]{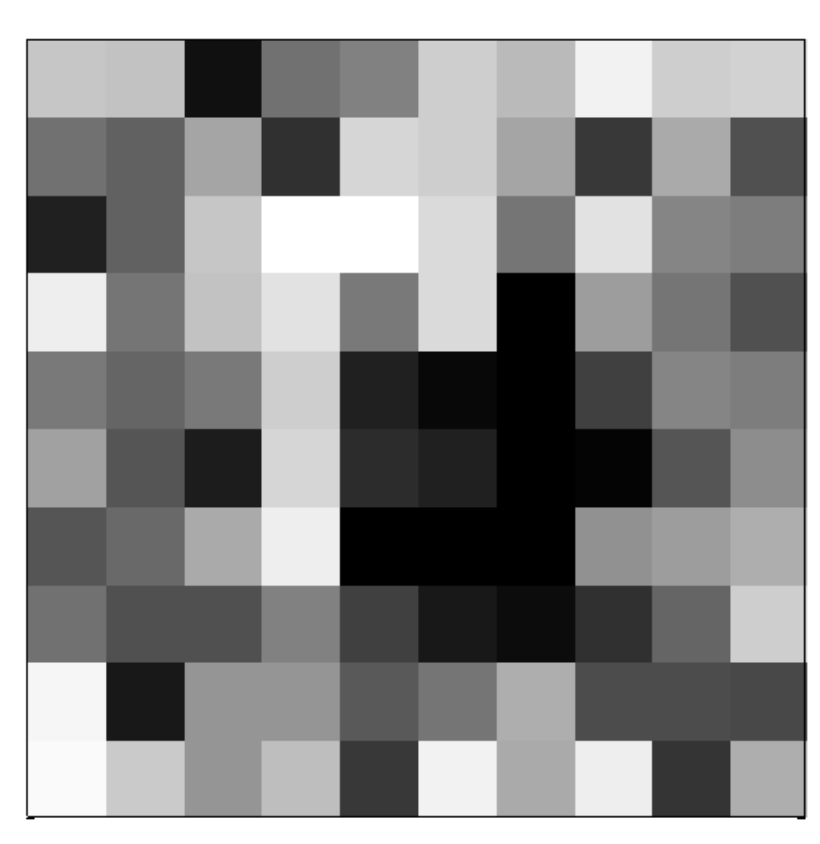}
\end{minipage}\hspace{-0.2em}
\begin{minipage}{0.035\textwidth}
\includegraphics[width=0.95\textwidth]{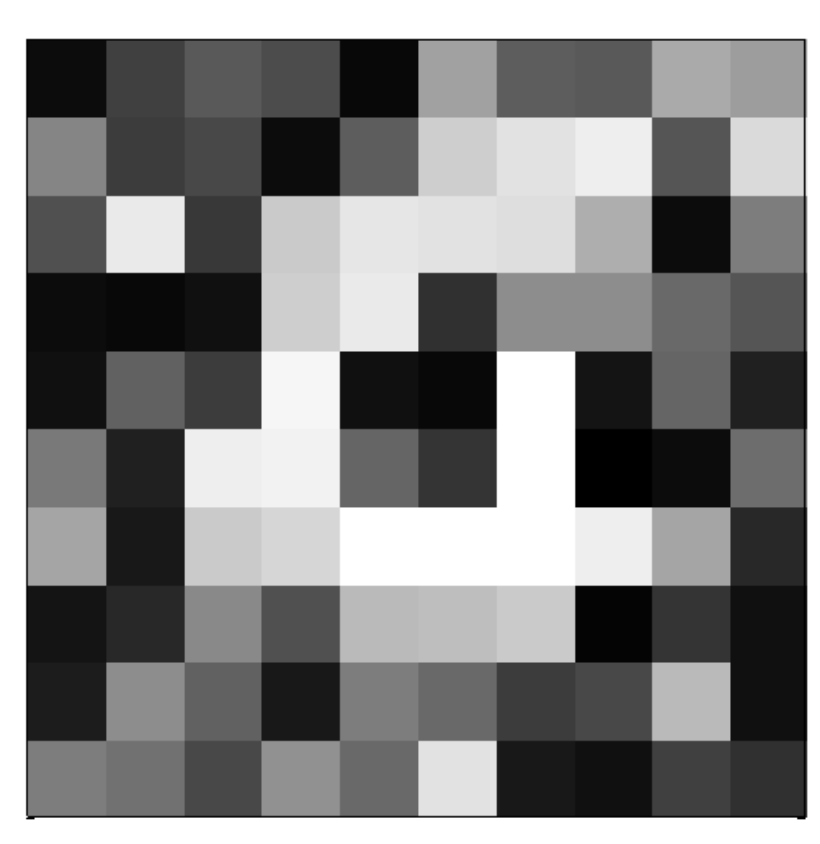}
\end{minipage}\hspace{-0.2em} 
\\
    \rotatebox[origin=Bc]{0}{\parbox{2cm}{{NGFA}}}
    &
\begin{minipage}{0.035\textwidth}
\includegraphics[width=0.95\textwidth]{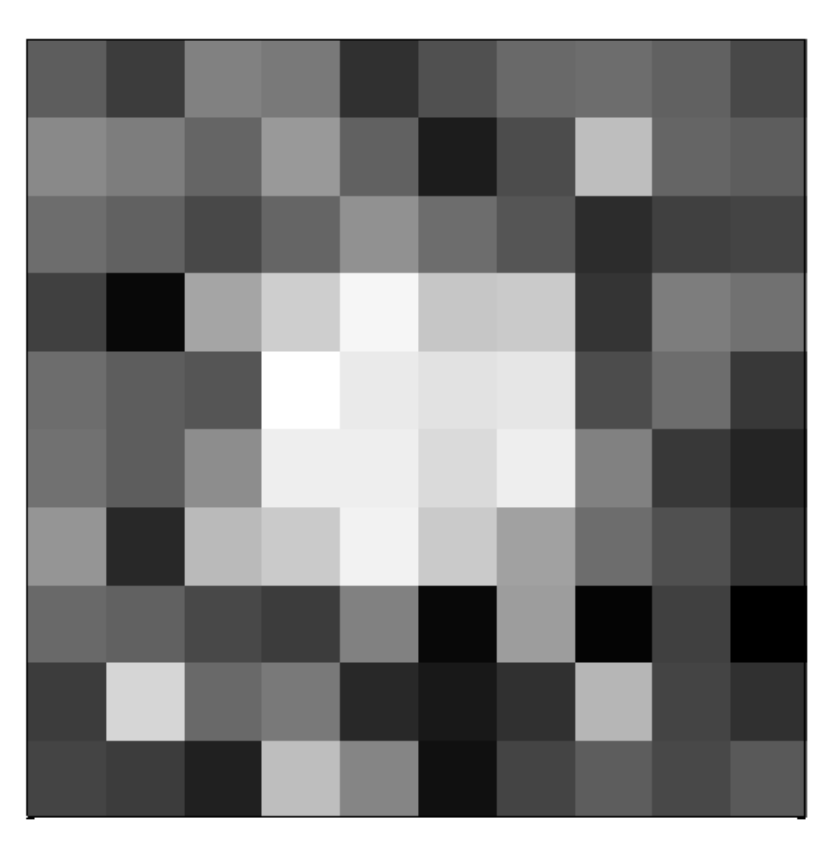}
\end{minipage}\hspace{-0.2em}
\begin{minipage}{0.035\textwidth}
\includegraphics[width=0.95\textwidth]{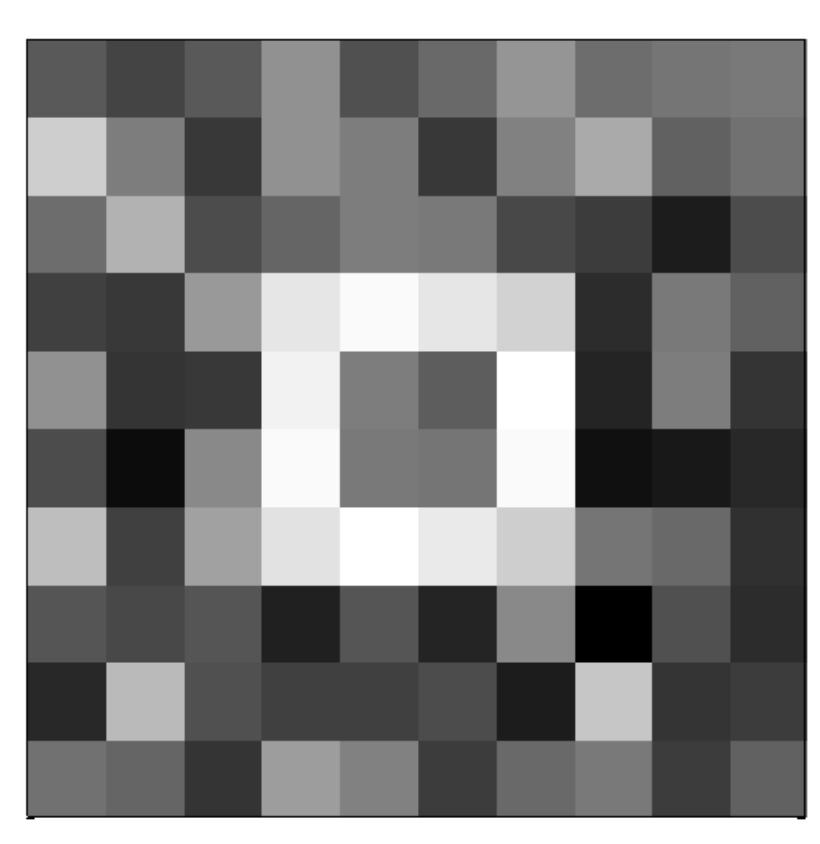}
\end{minipage}\hspace{-0.2em}
\begin{minipage}{0.035\textwidth}
\includegraphics[width=0.95\textwidth]{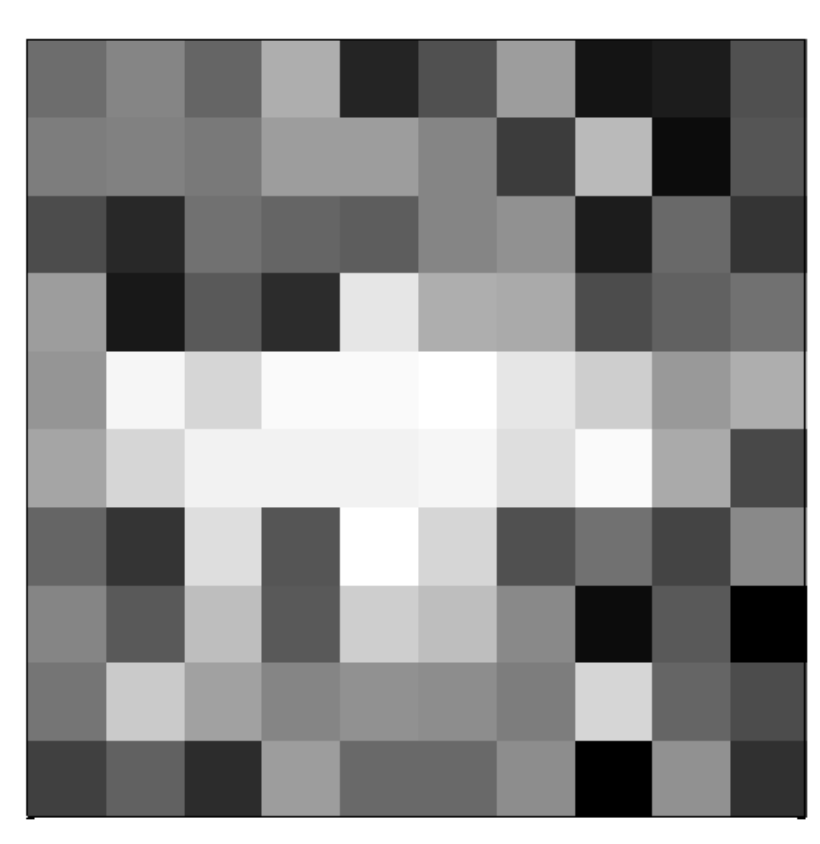}
\end{minipage}\hspace{-0.2em}
\begin{minipage}{0.035\textwidth}
\includegraphics[width=0.95\textwidth]{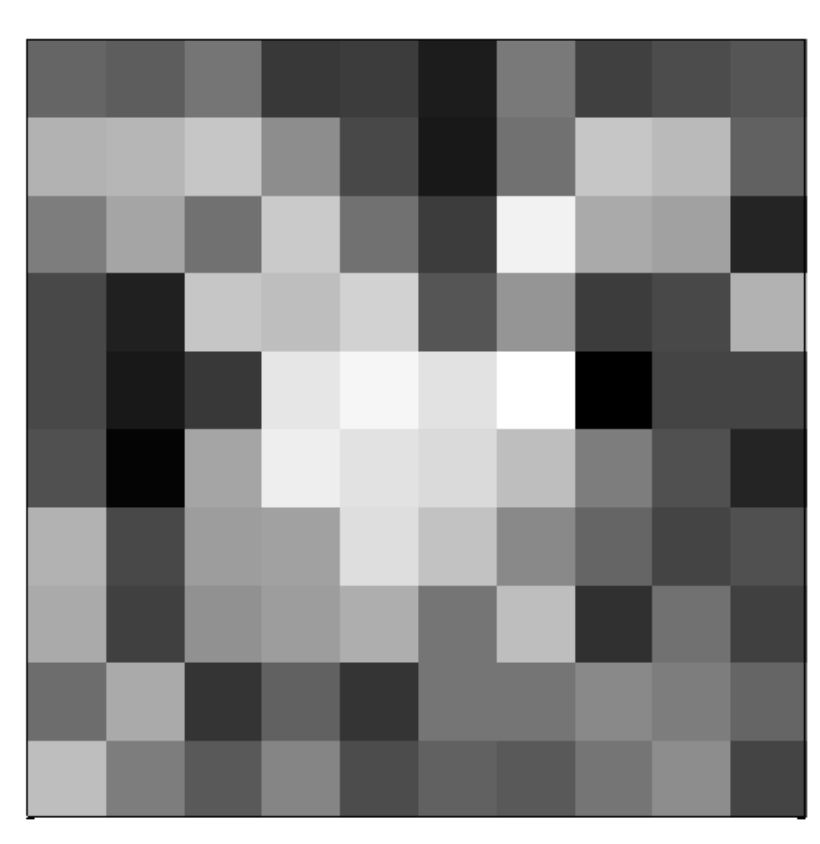}
\end{minipage}\hspace{-0.2em}
\begin{minipage}{0.035\textwidth}
\includegraphics[width=0.95\textwidth]{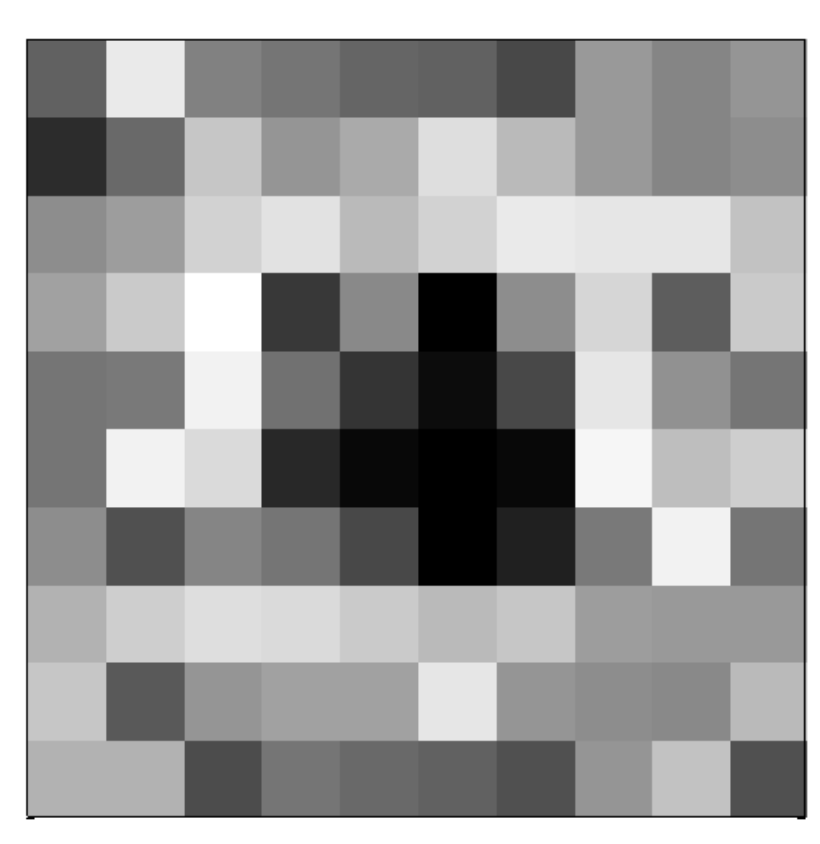}
\end{minipage}\hspace{-0.2em}
\begin{minipage}{0.035\textwidth}
\includegraphics[width=0.95\textwidth]{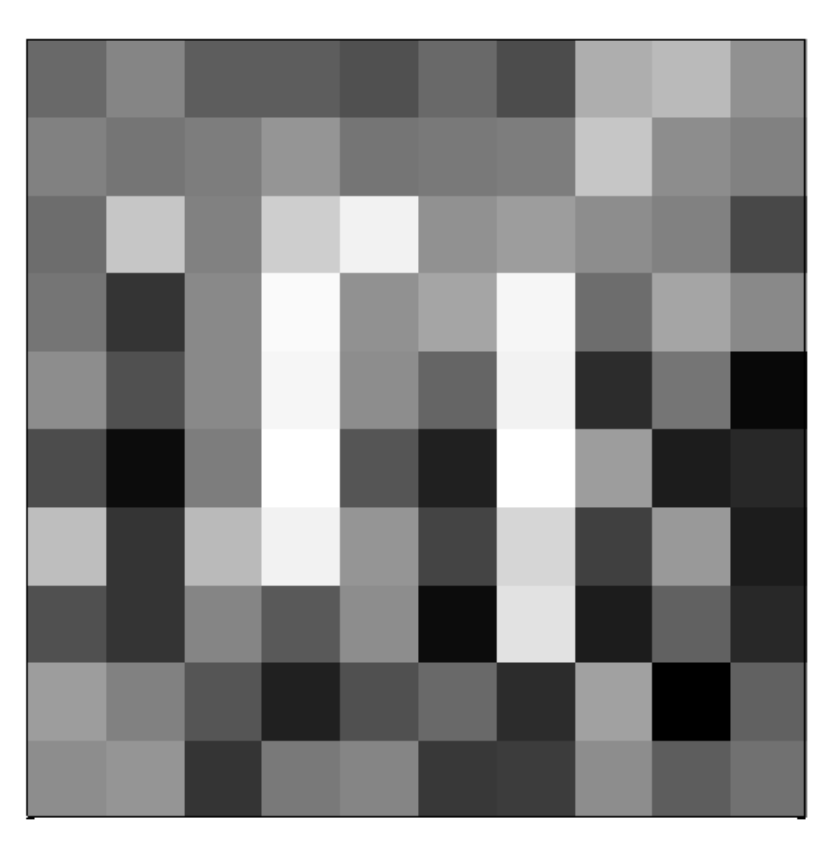}
\end{minipage}\hspace{-0.2em}
\begin{minipage}{0.035\textwidth}
\includegraphics[width=0.95\textwidth]{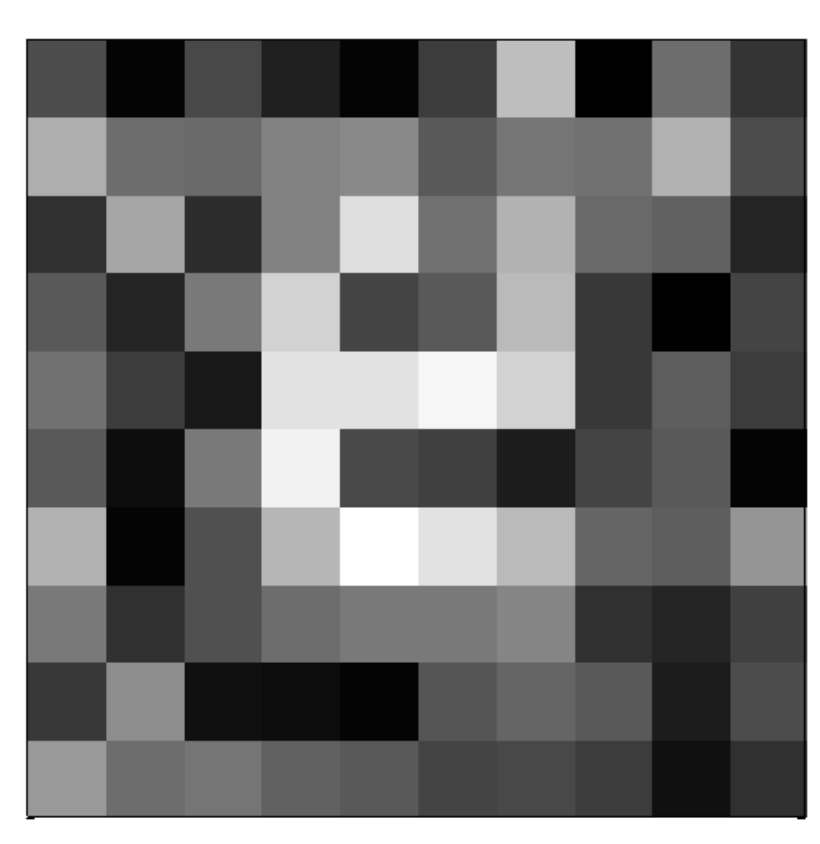}
\end{minipage}\hspace{-0.2em}
\begin{minipage}{0.035\textwidth}
\includegraphics[width=0.95\textwidth]{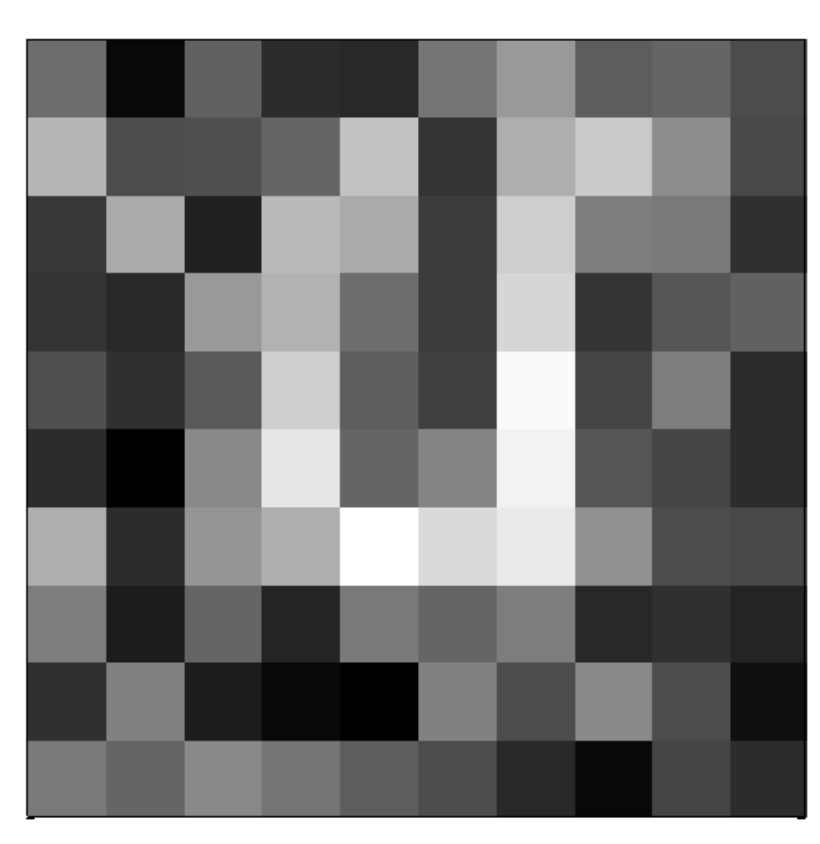}
\end{minipage}\hspace{-0.2em}
\begin{minipage}{0.035\textwidth}
\includegraphics[width=0.95\textwidth]{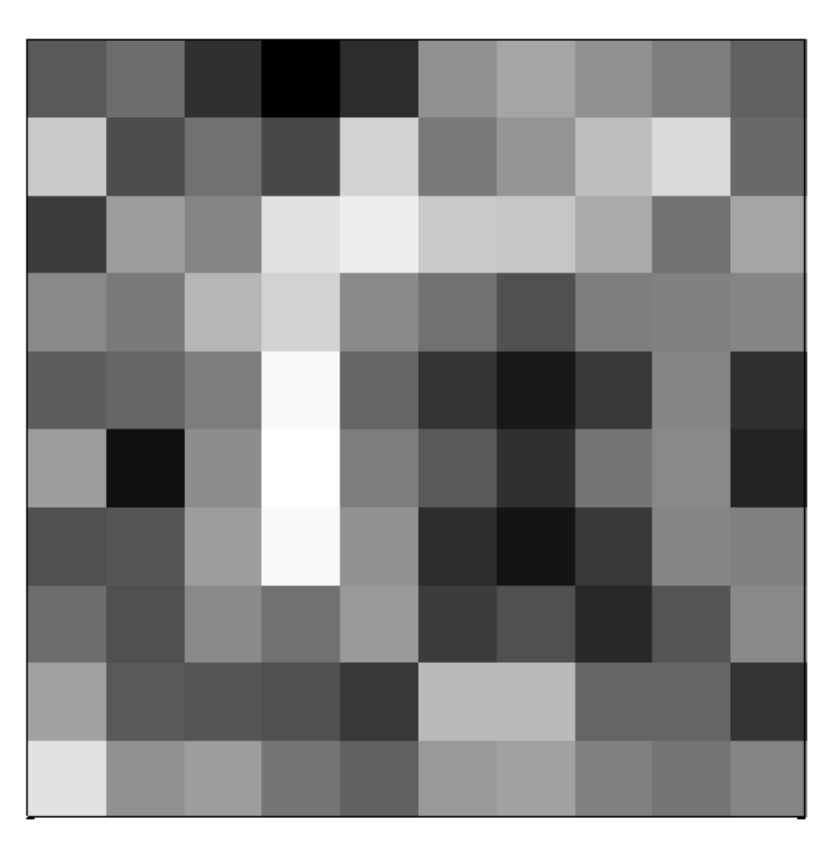}
\end{minipage}\hspace{-0.2em}
\begin{minipage}{0.035\textwidth}
\includegraphics[width=0.95\textwidth]{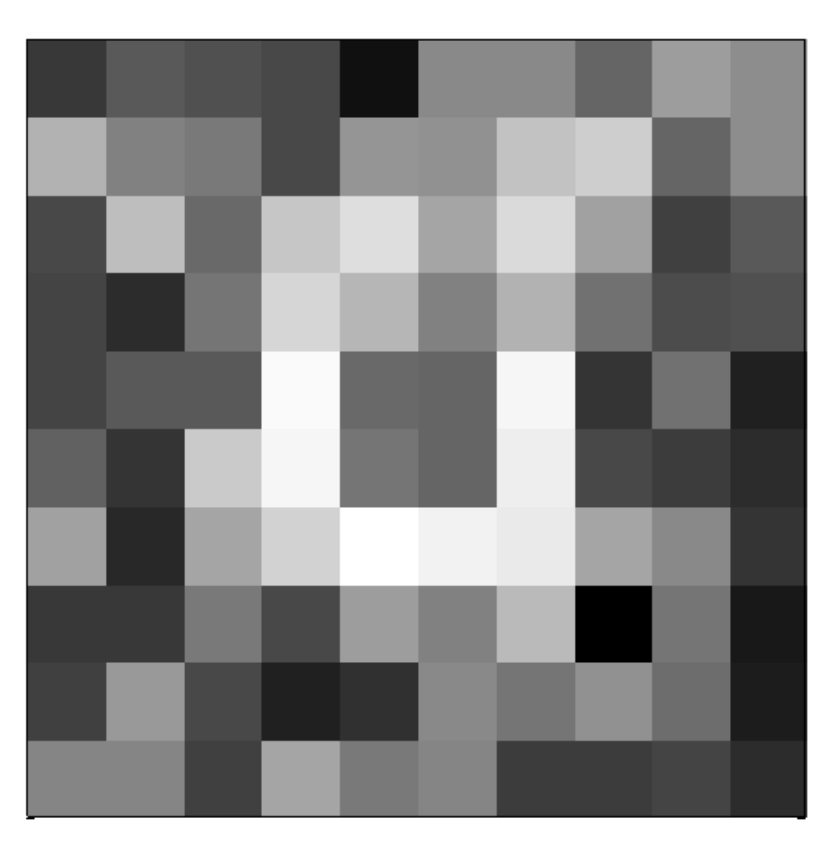}
\end{minipage}\hspace{-0.2em} 
  \end{tabular}
  \caption{Presented images (first row) and the reconstructed visual images obtained from the BCCA (second row) and the NGFA (third row).}
  \label{fMRI01}
\end{figure}

Bayesian canonical correlation analysis (BCCA) was investigated to analyze fMRI responses to visual stimuli in~\citep{BCCA_fMRI_01}. We evaluate the NGFA using our CVI algorithm to the fMRI recordings of two subjects viewing visual images consisting of contrast-defined $10\times10$ patches \citep{neuron}. The data is composed of two independent sessions: one for \enquote{random image session} with spatially random patterns sequentially presented; the other for \enquote{figure image session} with alphabet letters and geometric shapes sequentially presented. For the NGFA, we first treat the random image session and corresponding fMRI recordings as two groups, to extract the image bases and weight vector automatically from the input, with the initial $K$ set to $\min(D_1, D_2) = 100$. Our task is to reconstruct the visual image from the new fMRI recordings in the figure image session. The reconstruction performance is evaluated by the mean squared error between the presented and reconstructed images. We run both the BCCA and the NGFA for 20 times. The mean squared prediction error over 20 runs for NGFA is 0.224 with the standard deviation less than 1e-3, which is better than the result 0.251(0.002) of the BCCA. The reconstructed geometric shapes and alphabet letters by the BCCA and the proposed NGFA are shown in Fig.~\ref{fMRI01}.

\section{Discussion}
In this work, the GFA problem is tackled via a Bayesian nonparametric method that allows the total number of factors to be automatically inferred, and the underlying structured sparsity to be effectively captured.
In particular, we have presented an efficient collapsed variational inference algorithm for the nonparametric Bayesian group factor analysis model.
By integrating out the group-specific beta process parameters, our CVI algorithm achieves a better approximation because all latent variables are dependent through the field while the weak dependences are very small in the collapsed space. Using the Gaussian approximation technique, all the variational parameters can be efficiently updated through closed form expressions. Experimental results on both synthetic data and real-world applications demonstrate superior performance of our CVI algorithm for the nonparametric Bayesian group factor analysis model when compared to state-of-the-art GFA methods. 
An interesting direction of future research is how to infer hierarchically structured latent factors, as was done for deep factor modelling~\citep{DSBM, DPFA, AGBN}. Another possible direction would be to generalize GFA methods to model dynamic multiple related graph data~\citep{Durante} under the Poisson factorization framework~\citep{EPM,DPGM}.
\section*{Acknowledgements}
The authors thank the anonymous reviewers, Nurgazy Sulaimanov and Nikita Kruk for their useful comments
and suggestions.
This research is funded by the European Union's Horizon 2020 research
and innovation programme under grant agreement 668858.

\bibliographystyle{apa}
\bibliography{ref}

\begin{thebibliography}{}

\bibitem[\protect\astroncite{Blei et~al.}{2003}]{LDA}
Blei, D.~M. et~al. (2003).
\newblock Latent {D}irichlet allocation.
\newblock {\em JMLR}, 3:993--1022.

\bibitem[\protect\astroncite{Bunte et~al.}{2016}]{SGFA2016}
Bunte, K. et~al. (2016).
\newblock Sparse group factor analysis for biclustering of multiple data
  sources.
\newblock {\em Bioinformatics}.

\bibitem[\protect\astroncite{Carvalho et~al.}{2008}]{Carvarlho2008}
Carvalho, C.~M. et~al. (2008).
\newblock High-dimensional sparse factor modeling: Applications in gene
  expression genomics.
\newblock {\em JASA}, 103(484):1438--1456.

\bibitem[\protect\astroncite{Chen et~al.}{2011}]{Chen2011}
Chen, B. et~al. (2011).
\newblock The hierarchical beta process for convolutional factor analysis and
  deep learning.
\newblock In {\em ICML}, pages 361--368.

\bibitem[\protect\astroncite{Durante et~al.}{2017}]{Durante}
Durante, D. et~al. (2017).
\newblock {B}ayesian learning of dynamic multilayer networks.
\newblock {\em JMLR}, 18(1):1414--1442.

\bibitem[\protect\astroncite{Foulds et~al.}{2013}]{SCVBLDA}
Foulds, J. et~al. (2013).
\newblock Stochastic collapsed variational {B}ayesian inference for latent
  {D}irichlet allocation.
\newblock In {\em KDD}, pages 446--454.

\bibitem[\protect\astroncite{Fox et~al.}{2011}]{fox2011}
Fox, E.~B. et~al. (2011).
\newblock A sticky hdp-hmm with application to speaker diarization.
\newblock {\em Ann. Appl. Stat.}, 5:1020--1056.

\bibitem[\protect\astroncite{Fujiwara et~al.}{2009}]{BCCA_fMRI_01}
Fujiwara, Y. et~al. (2009).
\newblock Estimating image bases for visual image reconstruction from human
  rain activity.
\newblock In {\em NIPS}, pages 576--584.

\bibitem[\protect\astroncite{Gan et~al.}{2015a}]{DSBM}
Gan, Z. et~al. (2015a).
\newblock {Learning Deep Sigmoid Belief Networks with Data Augmentation}.
\newblock In {\em AISTATS}, pages 268--276.

\bibitem[\protect\astroncite{Gan et~al.}{2015b}]{DPFA}
Gan, Z. et~al. (2015b).
\newblock Scalable deep {P}oisson factor analysis for topic modeling.
\newblock In {\em ICML}, pages 1823--1832.

\bibitem[\protect\astroncite{Gupta et~al.}{2012a}]{GibbsHBP}
Gupta, S.~K. et~al. (2012a).
\newblock A {B}ayesian nonparametric joint factor model for learning shared and
  individual subspaces from multiple data sources.
\newblock In {\em SDM}, pages 200--211.

\bibitem[\protect\astroncite{Gupta et~al.}{2012b}]{SliceHBP}
Gupta, S.~K. et~al. (2012b).
\newblock A slice sampler for restricted hierarchical beta process with
  applications to shared subspace learning.
\newblock In {\em UAI}, pages 316--325.

\bibitem[\protect\astroncite{Hoef}{2012}]{delta_method}
Hoef, J. M.~V. (2012).
\newblock Who invented the delta method?
\newblock {\em The American Statistician}, 66(2):124--127.

\bibitem[\protect\astroncite{Ishiguro et~al.}{2017}]{ACVB}
Ishiguro, K. et~al. (2017).
\newblock Averaged collapsed variational {B}ayes inference.
\newblock {\em JMLR}, 18(1):1--29.

\bibitem[\protect\astroncite{Klami et~al.}{2013}]{BCCA2013}
Klami, A., Virtanen, S., and Kaski, S. (2013).
\newblock {B}ayesian canonical correlation analysis.
\newblock {\em JMLR}, 14(1):965--1003.

\bibitem[\protect\astroncite{Knowles et~al.}{2011}]{NSFA2011}
Knowles, D.~A. et~al. (2011).
\newblock Nonparametric {B}ayesian sparse factor models with application to
  gene expression modeling.
\newblock {\em Ann. Appl. Stat.}, 5(2B):1534--1552.

\bibitem[\protect\astroncite{Lahti et~al.}{2013}]{CGP2012}
Lahti, L. et~al. (2013).
\newblock Cancer gene prioritization by integrative analysis of m{RNA}
  expression and {DNA} copy number data.
\newblock {\em Briefings in Bioinformatics}, 14(1):27--35.

\bibitem[\protect\astroncite{Miyawaki et~al.}{2008}]{neuron}
Miyawaki, Y. et~al. (2008).
\newblock {Visual Image Reconstruction from Human Brain Activity using a
  Combination of Multiscale Local Image Decoders}.
\newblock {\em Neuron}, pages 5--29.

\bibitem[\protect\astroncite{Paisley et~al.}{2009}]{BPFA2009}
Paisley, J. et~al. (2009).
\newblock Nonparametric factor analysis with beta process priors.
\newblock In {\em ICML}, pages 777--784.

\bibitem[\protect\astroncite{Pitman}{2006}]{csp}
Pitman, J. (2006).
\newblock {\em Combinatorial stochastic processes}.
\newblock Springer-Verlag, Berlin.
\newblock Lectures on Probability Theory.

\bibitem[\protect\astroncite{Rai and {Daume III}}{2008}]{Piyushi2008}
Rai, P. and {Daume III}, H. (2008).
\newblock The infinite hierarchical factor regression model.
\newblock In {\em NIPS}, pages 1321--1328.

\bibitem[\protect\astroncite{Teh et~al.}{2006}]{CVBLDA}
Teh, Y.~W. et~al. (2006).
\newblock A collapsed variational {B}ayesian inference algorithm for latent
  {D}irichlet allocation.
\newblock In {\em NIPS}, pages 1353--1360.

\bibitem[\protect\astroncite{Teh et~al.}{2007}]{HDP}
Teh, Y.~W. et~al. (2007).
\newblock Hierarchical {D}irichlet processes.
\newblock {\em JASA}, 101(476):1566--1581.

\bibitem[\protect\astroncite{Teh et~al.}{2008}]{CVBHDP}
Teh, Y.~W. et~al. (2008).
\newblock Collapsed variational inference for {HDP}.
\newblock In {\em NIPS}, pages 1481--1488.

\bibitem[\protect\astroncite{Thibaux et~al.}{2007}]{HBP}
Thibaux, R. et~al. (2007).
\newblock Hierarchical beta processes and the {I}ndian buffet process.
\newblock In {\em AISTATS}, pages 564--571.

\bibitem[\protect\astroncite{Virtanen et~al.}{2012}]{BGFA2012}
Virtanen, S. et~al. (2012).
\newblock {B}ayesian group factor analysis.
\newblock In {\em AISTATS}, pages 1269--1277.

\bibitem[\protect\astroncite{Wainwright et~al.}{2008}]{GMEFVI2008}
Wainwright, M.~J. et~al. (2008).
\newblock Graphical models, exponential families, and variational inference.
\newblock {\em Found. Trends Mach. Learn.}

\bibitem[\protect\astroncite{Wang et~al.}{2013}]{CVBHMM}
Wang, P. et~al. (2013).
\newblock Collapsed variational {B}ayesian inference for hidden markov models.
\newblock In {\em AISTATS}, pages 599--607.

\bibitem[\protect\astroncite{West}{2003}]{West03}
West, M. (2003).
\newblock {B}ayesian factor regression models in the "large p, small n"
  paradigm.
\newblock In {\em Bayesian Statistics}, pages 723--732.

\bibitem[\protect\astroncite{Witten et~al.}{2009}]{PMA2009}
Witten, D.~M. et~al. (2009).
\newblock A penalized matrix decomposition, with applications to sparse
  principal components and canonical correlation analysis.
\newblock {\em Biostatistics}, 10(3):515--534.

\bibitem[\protect\astroncite{Yang and Koeppl}{2018}]{DPGM}
Yang, S. and Koeppl, H. (2018).
\newblock Dependent relational gamma process models for longitudinal networks.
\newblock In {\em ICML}, pages 5551--5560.

\bibitem[\protect\astroncite{Zhang et~al.}{2017}]{CVBMJP}
Zhang, B. et~al. (2017).
\newblock Collapsed variational bayes for {M}arkov jump processes.
\newblock In {\em NIPS}, pages 3749--3757.

\bibitem[\protect\astroncite{Zhao et~al.}{2016}]{BASS2016}
Zhao, S. et~al. (2016).
\newblock {B}ayesian group factor analysis with structured sparsity.
\newblock {\em JMLR}, 17(196):1--47.

\bibitem[\protect\astroncite{Zhou}{2015}]{EPM}
Zhou, M. (2015).
\newblock Infinite edge partition models for overlapping community detection
  and link prediction.
\newblock In {\em AISTATS}, pages 1135--1143.

\bibitem[\protect\astroncite{Zhou et~al.}{2015}]{NBP}
Zhou, M. et~al. (2015).
\newblock Negative binomial process count and mixture modeling.
\newblock {\em {IEEE} Trans. PAMI}, 37(2):307--320.

\bibitem[\protect\astroncite{Zhou et~al.}{2016}]{AGBN}
Zhou, M. et~al. (2016).
\newblock Augmentable gamma belief networks.
\newblock {\em JMLR}, 17:1--44.

\bibitem[\protect\astroncite{Zou et~al.}{2006}]{SPCA2006}
Zou, H. et~al. (2006).
\newblock Sparse principal component analysis.
\newblock {\em J. Comput. and Graph. Statist.}, 15(2):265--286.

\end{thebibliography}

\clearpage

\onecolumn
\section*{Appendix: Collapsed Variational Inference for Nonparametric Bayesian Group Factor Analysis}

\subsection*{A.2 Variational Approximation}
Here, we provide the full variational approximation used in our CVI algorithm for the NGFA.
We use ``$\cdot$'' as a index summation shorthand, e.g., $x_{\cdot j} = \sum_i x_{ij}$. 
We assume the variational posterior over the latent variables and parameters as
\begin{align}
&q(\mathbf{Z}, \mathbf{W}, \mathbf{F}, \bm{\beta}, \bm{\lambda}, \bm{\tau}, \bm{\alpha}, \mathbf{s}, \mathbf{t}, \bm{\eta}) \notag \\& = q(\mathbf{W}) q(\mathbf{F}) q(\bm{\beta}) q(\bm{\lambda}) q(\bm{\tau}) q(\bm{\alpha}) q(\mathbf{s}, \mathbf{t}, \bm{\eta} | \mathbf{Z}) q(\mathbf{Z}),\notag
\end{align}
where we define the variational posterior for each parameter as
\begin{align}
q(\mathbf{W}) & = \prod_{m, d, k} \mathcal{N}(w_{kd}^{\scriptscriptstyle(m)} ; \mu^{\scriptscriptstyle(m)}_{w_{kd}}, \sigma^{\scriptscriptstyle(m)}_{w_{kd}}) 
,\notag\\
q(\mathbf{F}) & = \prod_{n,k} \mathcal{N}({f}_{nk} ; {\mu}_{f_{nk}}, {\sigma}_{f_{nk}})
,\notag\\
q(\bm{\beta}) & = \prod_k \mathrm{Beta}(\beta_k ; a_k, b_k)
,\notag\\
q(\bm{\lambda}) & = \prod_{m, d, k}\mathrm{Gam}(\lambda_{kd}^{\scriptscriptstyle(m)} ; e_{kd}^{\scriptscriptstyle(m)}, f_{kd}^{\scriptscriptstyle(m)})
,\notag\\
q(\bm{\tau}) & = \prod_{m, n}\mathrm{Gam}(\tau_n^{\scriptscriptstyle(m)} ; g_n^{\scriptscriptstyle(m)}, h_n^{\scriptscriptstyle(m)} )
,\notag\\
q(\bm{\alpha}) & = \prod_{m} \mathrm{Gam}(\alpha^{\scriptscriptstyle(m)} ; c^{\scriptscriptstyle(m)}, d^{\scriptscriptstyle(m)}) 
,\notag\\
q(\mathbf{Z})& = \prod_{m,d,k}\mathrm{Bern}({z}_{kd}^{\scriptscriptstyle(m)} ; \rho_{kd}^{\scriptscriptstyle(m)}) 
,\notag\\
q(\mathbf{s} | \mathbf{Z}) & = \prod_{m,k} { \hat{n}_{mk} \brack s_{mk}} (\mathsf{G}[\alpha^{{(m)}}\beta_k])^{s_{mk}}
,\notag\\
q(\mathbf{t} | \mathbf{Z}) & = \prod_{m,k} { \tilde{n}_{mk} \brack t_{mk}} (\mathsf{G}[\alpha^{{(m)}}(1 - \beta_k)])^{t_{mk}}
,\notag\\
q(\bm{\eta} | \mathbf{Z}) & = \prod_{m} \mathrm{Beta}(\eta_m ; \mathsf{E}\left[ \alpha^{{(m)}}\right], D_m).\notag
\end{align}

\subsection*{A.3 Evidence Lower Bound (ELBO)}

The log marginal likelihood of data is lower bounded as

\begin{align}
&\log p(\mathbf{X}\mid \kappa_0)\geq \mathsf{E}\left[p(\mathbf{X}, \mathbf{Z}, \bm{\theta},\mathbf{s}, \mathbf{t}, \bm{\eta} \mid  \kappa_0)\right]  
 - \mathsf{E}\left[q(\mathbf{Z}, \bm{\theta},\mathbf{s}, \mathbf{t}, \bm{\eta})\right] %
\notag
\\
& =  \mathsf{E}_{q(\mathbf{\theta},\mathbf{Z})}\Bigg[ \mathsf{E}_{q(\mathbf{s},\mathbf{t},\bm{\eta} | \mathbf{Z})}\Big[ \log \frac{p(\mathbf{X}, \mathbf{Z}, \bm{\theta},\mathbf{s}, \mathbf{t}, \bm{\eta} \mid \kappa_0)}{q(\mathbf{s},\mathbf{t},\bm{\eta} \mid \mathbf{Z})}\Big] 
- \log q(\mathbf{\theta},\mathbf{Z}) \Bigg] \notag\\
&  = \mathsf{E}_{q(\bm{\theta},\mathbf{Z})} \left[ \log p(\mathbf{X}, \mathbf{Z}, \bm{\theta}\mid  \kappa_0) -q(\bm{\theta},\mathbf{Z}) \right], \label{full_bound}
\end{align}
where the second equality holds provided that $q(\mathbf{s},\mathbf{t},\bm{\eta} \ |\ \mathbf{Z})$ is set to its true posterior.

To derive the variational update for each parameter, we expand the ELBO for each term in Eq.~\ref{full_bound} as
\begin{align}
&\log p(\mathbf{X}\mid \kappa_0)
 \geq \ \mathsf{E}\left[ \log p(\mathbf{X} | \mathbf{W}, \mathbf{Z}, \mathbf{F}, \bm{\tau}) \right] \notag\\  &+ \mathsf{E}\left[ \log p(\mathbf{W}) \right] - \mathsf{E}\left[\log q(\mathbf{W}) \right] 
  +  \mathsf{E}\left[  \log p(\mathbf{Z})  \right] - \mathsf{E}\left[ \log q(\mathbf{Z}) \right] \notag\\  &+  \mathsf{E}\left[  \log p(\mathbf{F})  \right] - \mathsf{E}\left[ \log q(\mathbf{F}) \right] 
  + \mathsf{E}\left[  \log p(\bm{\lambda})  \right] - \mathsf{E}\left[ \log q(\bm{\lambda}) \right] \notag\\ &+  \mathsf{E}\left[  \log p(\bm{\tau})  \right] - \mathsf{E}\left[ \log q(\bm{\tau}) \right] 
  + \mathsf{E}\left[  \log p(\bm{\alpha})  \right] - \mathsf{E}\left[ \log q(\bm{\alpha}) \right] \notag\\  &+  \mathsf{E}\left[  \log p(\bm{\beta})  \right] - \mathsf{E}\left[ \log q(\bm{\beta}) \right].\label{fullELBO}
\end{align}
\subsection*{A.4 Variational Updates}
The variational updates for each parameter are obtained by taking the derivate of the ELBO in Eq.~\ref{fullELBO} w.r.t. each parameter and setting it to zero.
\\
\noindent\textbf{Updates for the sufficient statistics:}
\begin{align}
  & \mathsf{E}\left[ \hat{n}_{mk} \right] = \sum_{d} \rho_{kd}^{\scriptscriptstyle(m)},\qquad  \mathsf{E}\left[ \tilde{n}_{mk} \right] = \sum_{d} (1-\rho_{kd}^{\scriptscriptstyle(m)}),
 \notag\\
  &p_{\scalebox{0.8}{+}}\hat{n}_{mk})  = 1 - \exp\big({\sum_d \log [1 - \rho_{kd}^{\scriptscriptstyle(m)}]}\big),\notag\\
  &p_{\scalebox{0.8}{+}}( \tilde{n}_{mk})  = 1 - \exp\big({\sum_d \log [\rho_{kd}^{\scriptscriptstyle(m)}]}\big),
  \notag\\
&\mathsf{E}_{\scalebox{0.8}{+}}[ \hat{n}_{mk}]  = \frac{\mathsf{E}[ \hat{n}_{mk}]}{p_{\scalebox{0.8}{+}}(\hat{n}_{mk})},\qquad  \mathsf{E}_{\scalebox{0.8}{+}}[ \tilde{n}_{mk}]  = \frac{\mathsf{E}[ \tilde{n}_{mk}]}{p_{\scalebox{0.8}{+}}(\tilde{n}_{mk})},
\notag\\
    & \mathsf{V}\left[ \hat{n}_{mk} \right]  = \mathsf{V}\left[ \tilde{n}_{mk} \right]  = \sum_{d} (1 - \rho_{kd}^{\scriptscriptstyle(m)}) \rho_{kd}^{\scriptscriptstyle(m)},
  \notag\\
&\mathsf{V}_{\scalebox{0.8}{+}}[ \hat{n}_{mk}]  =  \frac{\mathsf{V}[ \hat{n}_{mk}]}{p_{\scalebox{0.8}{+}}(\hat{n}_{mk})}
,\qquad \mathsf{V}_{\scalebox{0.8}{+}}[ \tilde{n}_{mk}] =  \frac{\mathsf{V}[ \tilde{n}_{mk}]}{p_{\scalebox{0.8}{+}}(\tilde{n}_{mk})}.
\label{quantities}
\end{align}
\\ 
 \noindent\textbf{Updates for $\sigma^{\scriptscriptstyle(m)}_{w_{kd}}$ and $\mu^{\scriptscriptstyle(m)}_{w_{kd}}$:}
\begin{align}
\sigma^{\scriptscriptstyle(m)}_{w_{kd}} & = \left( \mathsf{E}\left[ \lambda_{kd}^{\scriptscriptstyle(m)}\right] + \mathsf{E}\left[ z_{kd}^{\scriptscriptstyle(m)}\right] \sum\nolimits_n \mathsf{E}\left[\tau_n^{\scriptscriptstyle(m)}\right] \mathsf{E}\left[f_{nk}^2  \right] \right)^{-1}, \label{sigma_w_kd}\\
\mu^{\scriptscriptstyle(m)}_{w_{kd}} & = \sigma^{\scriptscriptstyle(m)}_{w_{kd}} \left( \mathsf{E} \left[ {z}_{kd}^{\scriptscriptstyle(m)} \right] \sum\nolimits_n \mathsf{E}\left[ \tau_n^{\scriptscriptstyle(m)} \right] \mathsf{E}\left[ f_{nk} \right] \left.\tilde{x}_{nd}^{\scriptscriptstyle(m)}\right.^{\scriptscriptstyle-k} \right).
\label{mu_w_kd}
\end{align}
\\ 
 \noindent\textbf{Updates for the auxiliary variables $\mathbf{s, t}$:}
\begin{align}
&\mathsf{E}[s_{\scriptscriptstyle mk}] 
\approx \mathsf{G}[\alpha^{{(m)}}\beta_k] p_{\scalebox{0.8}{+}}(\hat{n}_{mk}) \big( \Psi\big( \mathsf{G}[\alpha^{{(m)}}\beta_k]+\mathsf{E}_{\scalebox{0.8}{+}}[ \hat{n}_{mk}] \big) 
\notag
\\ 
&-\Psi( \mathsf{G}[\alpha^{{(m)}}\beta_k] ) + \frac{{\mathsf{V}_{\scalebox{0.8}{+}}[ \hat{n}_{mk}] \Psi'( \mathsf{G}[\alpha^{{(m)}}\beta_k]+\mathsf{E}_{\scalebox{0.8}{+}}[ \hat{n}_{mk}]})}{2} \big) \notag
,\\[5pt]
& \mathsf{E}[t_{\scriptscriptstyle mk}] 
\approx\ \mathsf{G}[\alpha^{{(m)}}\bar{\beta}_k] p_{\scalebox{0.8}{+}}(\tilde{n}_{mk}) \big( \Psi\big( \mathsf{G}[\alpha^{{(m)}}\bar{\beta}]+\mathsf{E}_{\scalebox{0.8}{+}}[ \tilde{n}_{mk}] \big) 
\notag
\\ 
&-\Psi( \mathsf{G}[\alpha^{{(m)}}\beta_k] ) + \frac{{\mathsf{V}_{\scalebox{0.8}{+}}[ \tilde{n}_{mk}] \Psi'( \mathsf{G}[\alpha^{{(m)}}\bar{\beta}]+\mathsf{E}_{\scalebox{0.8}{+}}[ \tilde{n}_{mk}]})}{2} \big). \label{e_t}
\end{align}
\\ 
 \noindent\textbf{Updates for ${\sigma}_{f_{nk}}$ and ${\mu}_{f_{nk}}$:}
\begin{align}
{\sigma}_{f_{nk}} & = \left( \sum\nolimits_{m,d} \mathsf{E}\left[ \tau_n^{\scriptscriptstyle(m)} \right]  \mathsf{E}\left[ z_{kd}^{\scriptscriptstyle(m)}\right] \mathsf{E}\left[ \left(w_{kd}^{\scriptscriptstyle(m)} \right)^2\right] + 1\right)^{-1}, \label{sigma_f_kn}\\
{\mu}_{f_{nk}} & = {\sigma}_{f_{nk}} \left(  \sum\nolimits_{m,d} \mathsf{E}\left[ \tau_n^{\scriptscriptstyle(m)} \right] \mathsf{E}\left[ z_{kd}^{\scriptscriptstyle(m)}\right] \mathsf{E}\left[ w_{kd}^{\scriptscriptstyle(m)} \right] \left.\tilde{x}_{nd}^{\scriptscriptstyle(m)}\right.^{\scriptscriptstyle-k}\right).
\label{mu_f_kn}
\end{align}
\\ 
 \noindent\textbf{Updates for $a_k$ and $b_k$:}
\begin{align}
a_k & = \kappa_0/K + \mathsf{E}\left[ s_{\cdot k}\right]
,
b_k = \kappa_0(1-1/K) + \mathsf{E}\left[ t_{\cdot k}\right]. \label{b_k}
\end{align}
\\ 
 \noindent\textbf{Updates for $e_{kd}^{\scriptscriptstyle(m)}$ and $f_{kd}^{\scriptscriptstyle(m)}$:}
\begin{align}
e_{kd}^{\scriptscriptstyle(m)} & = e_0 + 1/2
,\qquad 
f_{kd}^{\scriptscriptstyle(m)} = f_0 + \left({\mathsf{E}\left[ \left(w_{kd}^{\scriptscriptstyle(m)} \right)^2\right]}\right)/2. \label{f_kd}
\end{align}
\\ 
 \noindent\textbf{Updates for $ g_n^{\scriptscriptstyle(m)}$ and $h_n^{\scriptscriptstyle(m)}$:}
\begin{align}
g_n^{\scriptscriptstyle(m)} &= g_0 + ({D_m})/{2}
, h_n^{\scriptscriptstyle(m)} = h_0 + \left({ \mathsf{E}\left[ \|{\mathbf{x}_n^{\scriptscriptstyle(m)}-\mathbf{G}^{\scriptscriptstyle(m)}\mathbf{f}_n}\|^2 \right] }\right)/{2}. \label{h_n}
\end{align}
\begin{algorithm}[t]
\SetKwInOut{Input}{Input}
\SetKwInOut{Output}{Output}
\Input{Data $\mathbf{X}$, Model $\log p(\mathbf{X},\mathbf{Z}, \bm{\theta},\mathbf{s}, \mathbf{t}, \bm{\eta})$, maximum iteraction $\mathcal{J}$, variational approximation $q(\mathbf{Z}, \bm{\theta},\mathbf{s}, \mathbf{t}, \bm{\eta} ; \mathbf{\Phi)}$, and hyper-parameter $\kappa_0$}
\Output{Variational parameters $\mathbf{\Phi}$\footnotemark}
Initialize $\mathbf{\Phi}$ randomly. \\
\For{iter $= 1:\mathcal{J}$}{
  \For{$k=1$ to $K_{+}$\footnotemark}{
  Update $a_k$, $b_k$ (Eq.~\ref{b_k})\\
  \For{$m = 1$ to $M$}{
  Update the sufficient statistics in (Eq.~\ref{quantities})\\
  Calculate $\mathsf{E}[s_{\scriptscriptstyle mk}]$, $\mathsf{E}[t_{\scriptscriptstyle mk}]$ (Eq.~\ref{e_t})\\
  \For{$d=1 $ to $ D_m$}{
  Update $\rho_{kd}^{{(m)}}$ (Eq.~\ref{rho_kd}) in Fig.~\ref{eq_5}\\
  Update $\sigma^{{(m)}}_{w_{kd}}$, $\mu^{{(m)}}_{w_{kd}}$ (Eq.~\ref{sigma_w_kd};~\ref{mu_w_kd})\\
  Update $e_{kd}^{{(m)}}$ and $f_{kd}^{{(m)}} $ (Eq.~\ref{f_kd})\\
  }
  }
  \For{$n=1$ to $N$}{
  Update ${\sigma}_{f_{kn}}$ and ${\mu}_{f_{kn}}$ (Eq.~\ref{sigma_f_kn};~\ref{mu_f_kn})\\
  }
  }
  \For{$m = 1$ to $M$}{
  Update $c^{{(m)}}$ and $d^{{(m)}}$ (Eq.~\ref{d_m})\\
  Calculate $\mathsf{E}[\log \eta_m]$ (Eq.~\ref{eta})\\
  \For{$n = 1$ to $N$}{
  Update $g_n^{{(m)}}$ and $h_n^{{(m)}}$ (Eq.~\ref{h_n})\\
  }
  }
 }
 \caption{Collapsed variational inference for the NGFA}
 \label{alg}
\end{algorithm}
\footnotetext{For the sake of clarity, we use $\mathbf{\Phi}$ to denote all the variational parameters.}
\footnotetext{We use $K_{+}$ to denote the number of active factors as the hierarchical beta Bernoulli prior can shrink the coefficients of the redundant factors to zeros.}
 \noindent\textbf{Updates for $ c^{\scriptscriptstyle(m)}$ and $d^{\scriptscriptstyle(m)}$:}
\begin{align}
c^{\scriptscriptstyle(m)} & = c_0 + \mathsf{E}\left[ s_{m\cdot}\right] + \mathsf{E}\left[ t_{m\cdot}\right]
, d^{\scriptscriptstyle(m)} = d_0 - \mathsf{E}\left[ \log \eta_m\right]. \label{d_m} 
\end{align}
\\ 
 \noindent\textbf{Updates for the auxiliary variables $\eta$:}
\begin{align}
& \mathsf{E}[\log \eta_m] = \Psi(\mathsf{E}[\alpha^{{(m)}}]) - \Psi(\mathsf{E}[\alpha^{{(m)}}] + D_m).
\label{eta}
\end{align}
Altogether, our {CVI} algorithm for the NGFA is summarized in Algorithm \ref{alg}.

\end{document}